\documentclass[lettersize,journal]{IEEEtran}
\usepackage{amsmath,amsfonts}
\usepackage{algorithmic}
\usepackage{algorithm}
\usepackage{array}
\usepackage[caption=false,font=normalsize,labelfont=sf,textfont=sf]{subfig}
\usepackage{textcomp}
\usepackage{stfloats}
\usepackage{url}
\usepackage{verbatim}
\usepackage{graphicx}
\usepackage{cite}
\usepackage{fancyhdr}
\usepackage{xcolor}
\usepackage{pifont}

\usepackage{hyperref}
\hypersetup{hypertex=true,
            colorlinks=true,
            linkcolor=black,
            anchorcolor=black,
            citecolor=black}

\begin{document}

\title{Seeing your sleep stage: cross-modal distillation from EEG to infrared video}

\author{Jianan Han, Shaoxing Zhang, Aidong Men, Yang Liu, Ziming Yao, Yan Yan, Qingchao Chen\textsuperscript{\ding{41}}

\thanks{\ding{41} Corresponding author.}
\thanks{This work is supported by Peking University Medicine Seed Fund for Interdisciplinary Research (BMU2022MX011), Hygiene and Health Development Scientific Research Fostering Plan of Haidian District Beijing(HP2021-11-50102), the Fundamental Research Funds for the Central Universities, and PKU-OPPO Innovation Fund BO202103.}
}

\markboth{Journal of \LaTeX\ Class Files,~Vol.~14, No.~8, August~2021}%
{Shell \MakeLowercase{\textit{et al.}}: A Sample Article Using IEEEtran.cls for IEEE Journals}


\maketitle

\begin{abstract}
It is inevitably crucial to classify sleep stage for the diagnosis of various diseases. However, existing automated diagnosis methods mostly adopt the ``gold-standard'' Electroencephalogram (EEG) or other uni-modal sensing signal of the PolySomnoGraphy (PSG) machine \textit{in hospital}, that are expensive, importable and therefore unsuitable for point-of-care monitoring \textit{at home}.
To enable the sleep stage monitoring \textit{at home}, in this paper, we analyze the relationship between infrared videos and the EEG signal and propose a new task: to classify the sleep stage using infrared videos by distilling useful knowledge from EEG signals to the visual ones. It is different from previous video classification and multi-modal analysis tasks, mainly in that (i) the temporal duration of the infrared video is relatively long (10 hours per night); (ii) and the semantic gap between the EEG and infrared video is disparate and much larger than conventional cross-modal data in multimedia analysis such as video and audio. To establish a solid cross-modal benchmark for this application, we develop a new dataset termed as Seeing your Sleep Stage via Infrared Video and EEG ($S^3VE$). $S^3VE$ is a large-scale dataset including synchronized infrared video and EEG signal for sleep stage classification, including 105 subjects and 154,573 video clips that is more than 1100 hours long. Our contributions are not limited to datasets but also about a novel cross-modal distillation baseline model namely the structure-aware contrastive distillation (SACD) to distill the EEG knowledge to infrared video features. The SACD achieved the state-of-the-art performances on both our $S^3VE$ and the existing cross-modal distillation benchmark. Both the benchmark and the baseline methods will be released to the community. We expect to raise more attentions and promote more developments in the sleep stage classification and more importantly the cross-modal distillation from clinical signal/media to the conventional media. Code and open datasets are available at \textcolor{red}{https://github.com/SPIResearch/SACD}.
\end{abstract}


\section{Introduction}
\begin{figure*}
  \includegraphics[width=\textwidth]{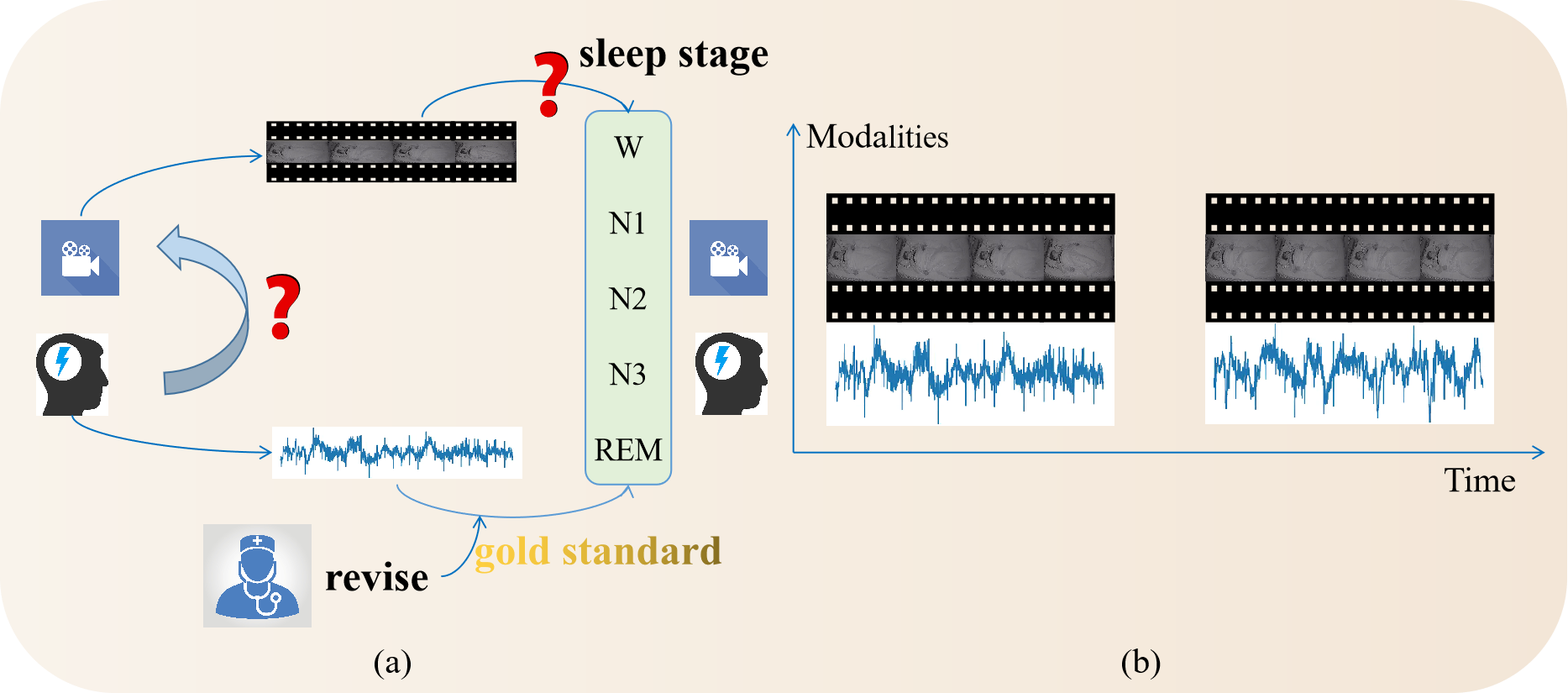}
  \caption{ (a). diagram of the task: cross-modal distillation from EEG to
infrared video (b). semantic gap between video and EEG modality. The example in the figure is selected from clips with both labels $N1$; the EEG signal can be successfully judged, but the second term is judged as $W$ through the video signal; In this $N1$ sleep state , some unknowing short-term movements affect the judgment of the IR video. For this example, the video cannot ``see'' the similarity of the two EEG signals, and the small movements of the legs and hands are equivalent to the confounding factor of the IR video modality.}
  \label{fig:mainteaser}
\end{figure*}

\IEEEPARstart{I}{t} is of significance to estimate the sleep quality and stage accurately, as it is directly related to the phenomenon (phenotype) of chronic disease and mental disease. According to the recent scientific research results, millions of chronic and mental disease patients have sleep related problems that are highly correlated to daily life dis-functioning and even traffic accidents etc. It is essential to tackle this global healthcare problems by measuring the sleep quality accurately and in-time especially at home.

Existing analysis methods and sleep quality or stage classification approaches adopt the usage of EEG signal as the ``gold standard'' sensing modality; however, it is time-consuming and costly to estimate the quality in the hospital via analyzing and diagnosing the EEG signal from the PSG machine. In addition, the annotation efforts and training of clinical workers to use and diagnosis are expensive. Moreover, it is nearly impossible to setup PSG machine at home as it is expensive and difficult to operate on. Besides, the PSG operation mode needs to attach tens of sensors on the patients' head and they sometimes fell off, which generates inaccurate results. \textit{\textbf{Therefore, it is of huge demand and extremely essential to estimate the sleep quality/stage using the portable and point-of-care sensors and solutions at home.}}

In this paper, we propose a novel cross-modal methodology to solve the previous barriers, enabling point-of-care sleep stage monitoring at home. We propose to sense the human body visually via an infra-red camera video synchronized with the PSG EEG signal. As shown in the Fig. \ref{fig:mainteaser} (a), with the help of EEG signal features and distilling EEG knowledge to the visual features, we investigate the possibility, capability and limitations of distilled infra-red visual features to classify the sleep stage. 

To enable the developments of point-of-care healthcare research and distillation methods from clinical to visual modality, to our best knowledge, we are the first to collect a large-scale cross-modal distillation dataset, namely $S^3VE$, including in total 1,100 hours synchronized infra-red videos and EEG signals, 105 subejcts from the real-world hospital and 154,573 multi-modal clips to investigate the problem of sleep stage classification. Besides the datasets, we also raise and analyze the following challenges of cross-modal distillation between the EEG and the infra-red data.

\textbf{Challenge 1: the large cross-modal semantic gap between EEG and IR signals.} EEG signal represents the intrinsic features of the sleep stage and is regarded as the gold-standard clinical diagnosis modality. However, the IR videos are commonly used for monitoring external characteristics of the subjects, e.g. motion, events and abnormal behaviour of the subjects. The two synchronized data in our $S^3VE$ are less correlated and with a large semantic gap than those in the conventional audio-video classification benchmark, e.g. UCF51. ( please see Fig. \ref{fig:mainteaser} (b) for more details). Therefore, we argue that directly aligning the cross-modal features of the same \textit{instance} may lead to inferior distillation performance and the collapsed joint embedding space.

\textbf{Challenge 2: the appearances of the infra-red sleep videos are \textbf{similar globally} and the differences among different classes' videos are \textbf{subtle locally}}. As shown in Fig. \ref{fig:mainteaser} (b), the global similarity between inter-class visual features are caused by the confounded similar background and scene. How to design a loss function to reflect and reveal the fine-grained \textit{contrast} between visual pairs of the same or different classes remains a challenge in our dataset. 

To tackle the first challenge, \textbf{\textit{we propose to align the relationship structure formed by multiple cross-modal clip features instead of aligning the individual instance}}. Due to the large semantic gap in our scenario, directly pulling the cross-modal features of the same \textit{instance} may lead to inferior distillation performance and the collapsed joint embedding space. Therefore, we alternates to align the structural information. The intuition is that in spite of the instance-level semantic gap, the uni-modal structure relationships should be similar among multiple instances as a group. In addition, the cross-modal data exhibit unique temporal characteristics and a strong temporal correlation among the features in a mini-batch is observed. Formulating both points as a constraint, we propose to build a graph among multiple instances in each unimodal data and regularize the consistency of two unimodal relationship structures.

To tackle the second challenge, \textbf{\textit{we propose to use the contrastive learning framework to learn the fine-grained ``contrast'' from the subtle differences in the IF video.}} However, conventional cross-modal contrastive learning is limited in our scenario, because the negative pairs are normally selected from the cross-modal features of different instances, which have large semantic gaps. Therefore, the normal negative pairs are not ``hard'' enough which cannot reflect and reveal the fine-grained contrast required in our setup. Therefore, we proposed to design two K-hard negative memory bank for the EEG and visual modality respectively, selecting the hard negative sample set in the online manner. In addition, we propose to use the symmetric contrastive cross-modal distillation to reduce the cross-modal semantic gap. 

To sum up, we made the following contributions in this paper: 

\begin{itemize}
    \item To our best knowledge, we proposed a new and the first dataset and benchmark to investigate cross-modal distillation between the clinical EEG signal and the IF videos. We provided extensive experiments and comparisons with conventional distillation methods in the dataset. 
    \item We also propose a novel cross-distillation method termed as Structure-aware contrastive distillation (SACD), including the structure-aware cross-modal alignment module and the dual memory banks for the contrastive learning.
    \item Our method achieves SOTA results on both our benchmark and the conventional distillation benchmark, e.g. UCF51. 
\end{itemize}

\section{Related works}
   
\subsection{Various Modalities of Sleep Stage Classification}
Accurate sleep stage classification has been of great interest in analysing sleep quality and determining the effectiveness of treatment. As the EEG signal is considered as the ``gold-standard'' for sleep stage classification, The most mainstream approach is classifying and analyzing sleep stage by employing physiological electrical signals. In addition to PSG-based sleep grading and apnea-related studies, there are many other approaches. For example, Goederen \textit{et al.}  \cite{de2021radar} used broadband radar to analyze children's sleep stages and sleep status. Deng \textit{et al.} \cite{deng2017decision} designed adaptive vertical box (AV-Box) based breathing/snoring detection using a decision tree classifier for sleep stage classification. Korkalainen \textit{et al.} \cite{korkalainen2020deep} identified the sleep stages from the photoplethysmogram (PPG) signal obtained with a simple finger pulse. Yi \textit{et al.} \cite{yi2019non} extracted a total of 74 features, including heart rate variability (HRV), features, respiratory rate variability (RV) features, and linear frequency cepstral coefficients (LFCC) from bed sensor data and performs sleep stage classification. \textit{\textbf{Unlike the above methods, we are the first to systematically investigate the use of infrared sleep video for sleep stage classification and have developed a new dataset.
       }}

\subsection{PSG-based Sleep Stage Classification Datasets and Methods} 

\noindent\underline{\textbf{Sleep Stage Classification Datasets:}} Various physio-electrical signal datasets have been collected for sleep research. \textit{The Sleep Heart Health Study (SHSS)} was a multi-center cohort sleep study \cite{quan1997sleep}\cite{zhang2018national}, whose two dataset versions, SHHS-1 and SHHS-2, contained the polysomnograms (PSG) data of 6441 and 3295 subjects respectively. The PSG data consist of multi-channel physio-electrical signals, including the Electroencephalogram(EEG) (C3-A2 and C4-A1), Electrooculogram (EOG), Electromyogram (EMG), Thoracic excursions (THOR) and abdominal excursions (ABDO), etc. \textit{The SleepEDF-20 and SleepEDF-78} were obtained from the PhysioBank \cite{goldberger2000physiobank}, including 20 and 78 subjects respectively. The data contains 2 EEG channels (Fpz-Cz and Pz-Oz). \textit{The Montreal Archive of Sleep Studies (MASS) dataset} \cite{o2014montreal} was collected including the whole-night sleep data from 200 subjects (103 females and 97 males), aged from 18-76. It mainly consists of about 20 EEG channels, plus EOG, EMG, ECG, and respiration signals. \textit{\textbf{Different from the previous datasets, to our best knowledge, we proposed a novel cross-modal dataset to promote multi-modal learning for sleep stage classification, including the synchronous IR videos and the EEG signals.}}

\noindent\underline{\textbf{PSG-based Automated Sleep Stage Classification Methods:}} Multiple sleep classification methods have been proposed using the previously mentioned datasets. Conventional machine learning methods extracted time-frequency analysis features and applied the Support Vector Machine (SVM), random forest, wavelet transform and information entropy algorithms \cite{fraiwan2010classification}\cite{jiang2019robust}\cite{liang2012automatic}. However, these methods incorporated strong prior knowledge and hand-crafted features, therefore the classification accuracy relies on the feature qualities. Recently, deep learning based methods have been the mainstream for sleep stage classification. For single-channel EEG classification, the AttnSleep \cite{eldele2021attention} used the multi-resolution convolutional neural network (MRCNN). The dilated convolution and synthetic minority oversampling technique (SMOTE)  \cite{jia2020sleep} also achieved competitive results. As for methods using multi-channel EEG signals, the BrainSleepNet \cite{cai2020brainsleepnet} captured the comprehensive features of multi-channel EEG signals. The MSTGCN \cite{jia2021multi} and GraphSleepNet \cite{jia2020graphsleepnet} used the structure-aware encoders for automatic sleep staging. In addition, a joint CNN framework \cite{phan2018joint} adopted temporal information as a context to predict sleep stages. \textbf{\textit{However, our dataset consists of sleep data of 105 subjects and most differently, we leveraged the synchrounous EEG and infra-red videos to analyze and trian the classification model, which is not available in the above datasets. To our best knowledge, we are the first to investigate the usage of EEG signals and video for sleep stages classification.}}



\subsection{Cross-modal distillation methods}
 
\noindent\underline{\textbf{Knowledge distillation (KD):}} The KD \cite{hinton2015distilling} \cite{ba2014deep}\cite{bucila2006model} \cite{girdhar2019distinit} methods transfer knowledge from the teacher to the student network, by supervising the student network using the pseudo labels. Komodakis \textit{et al.}  \cite{komodakis2017paying} used an attention-based distillation method to match the activation-based and gradient-based spatial attention maps. The flow of solution procedure (FSP) \cite{yim2017gift}, generated by computing the Gram matrix of features across layers, was used to transfer knowledge. The RKD \cite{park2019relational} captures cross-instance relations by designing a loss function to penalize the structure variations. Li \textit{et al.} \cite{li2017learning} developed a new framework to correct noisy labels by using knowledge learned from small clean datasets and semantic knowledge graphs. \textbf{\textit{Different from them, we not only pull the features between the student and the teacher directly, but also allows the student network to learn the relative positional relationships between instances in the teacher's network during the distillation process. This relative positional relationship, specifically (which instances are close, far, and how they are distributed), within a mini-batch, is what we call ``structure.'' Hence the name of our distillation method is ``structure-aware.'' }}

\noindent\underline{\textbf{Structure—aware distillation methods:}}
Structure-aware distillation adopts the idea of distilling knowledge from structural data. The CRCD \cite{tung2019similarity} estimates the mutual relation and transfers structured knowledge from anchor teacher to anchor student in a contrastive learning framework. Pairwise distillation methods distilled pairwise and holistic similarities \cite{liu2019structured}. The similarity-preserving KD \cite{zhu2021complementary} constrains the similarity between teacher network features and the student one, which complements the conventional distillation methods.
\textbf{\textit{However, different from their works, we propose to a new method use graph neural network to model the similarity relationship between different modalities, by updating the  ``node'' and  ``edge'' inside one modality, forming an entire graph-level representation to describe this modality and then take them closer.}}

\begin{table*}[t]
\small
\centering
\caption{Characteristics of five sleep stages.}
\setlength\tabcolsep{0.55pt}
\resizebox{0.8\linewidth}{!}{
\begin{tabular}{c|l}
\hline
\multicolumn{1}{l|}{Sleep stages} & \multicolumn{1}{c}{Characteristics}                                                                      \\ \hline
$W$                                 & Awake. An EEG contains $\beta$ waves when the eyes are closed and $\alpha$ waves when the eyes are open. \\ \hline
$N1$                                & Transitions from $W$ to other stages. Cranial apex waves are present in later stages.                      \\ \hline
$N2$                                & Spindles or unawakening associated $K$-complex waves are present.                                          \\ \hline
$N3$                                & High amplitude low frequency $\sigma$ wave appears.                                                      \\ \hline
$REM$                               & There is rapid eye movement and typical sawtooth wave                                                    \\ \hline
\end{tabular}}
\label{tab: stage}
\end{table*}

\noindent \underline{\textbf{Cross-modal distillation using contrastive learning:}}
Most recently, KD methods using the contrastive learning pulls the representations of positive pairs but push the negative pairs. The Contrastive Multiview Coding (CMC) \cite{tian2020contrastive} is a cross-view learning method to align different views of the same instances by contrastive learning. The Contrastive Representation Distillation (CRD) \cite{tian2019contrastive} transfers knowledge by instance-level contrastive learning and uses a large memory bank to store negative samples.
Chen et al. \cite{chen2021distilling} proposed to use contrastive learning to distill information from the image and audio to video analysis. \textit{\textbf{Different from other contrastive learning KD method, we proposed a novel dual-modality K-hard negative queues, storing the negative samples of the sleep stages. In addition, we designed the symmetric contrastive distillation losses, leveraging negative pairs from two modalities.}} 

\section{Datasets and Benchmarks}
\subsection{Problem Formulation}
Sleep disordering has become an important problem and it is of great significance to establish a sleep stage classification standard for sleep medicine. The American Academy of Sleep Medicine (AASM) standard\cite{berry2012aasm} sets out the rules, terminology and techniques for sleep and related events.
AASM divides human sleep into five stages, including the Rapid Eye Movements($REM$), Wake($W$), Non REM1 ($N1$), Non REM2 ($N2$) and Non REM3 ($N3$). The characteristics of each stage are shown in the following Table \ref{tab: stage} \cite{jia2021salientsleepnet}.

\subsection{Datasets and benchmarks construction}
\label{sec:3.2}
\noindent\textbf{Data Collection.}
We collected the synchronized EEG and the infra-red video signals from the Peking University Third Hospital. The dataset collection time spans more than two years
In order to ensure the diversity of subjects in the dataset, 
we selected subjects of different ages (the youngest is 7 years old; the oldest is 70 years old), genders, and sleep apnea indices. Through consulting with a clinical expert, we formulated the training, validation, and test sets accordingly. The data collection time is approximately from 9:30pm to 7:30am the next day. 


\noindent\textbf{Annotations.} In the process of constructing the dataset annotations, the coarse-grained annotations are firstly generated from the PSG machine, and then they are inspected and examined by five well-trained sleep apnea physicians. In this process, some unreasonable and mis-classified labels are modified through discussions to achieve agreements. To eliminate the effects of subjective diagnosis, weekly cross-checks will be performed to ensure the agreed consensus. Conflicted annotations and opinions are collectively discussed and voted so that the annotation agreements within each subject should be more than 95\%. At present, the above annotation protocol is widely recognized by various medical institutions and hospitals, and is considered to be the ``gold-standard'' for sleep stage classification.

\subsection{Dataset Statistics and Properties}
\begin{figure}
  \includegraphics[width=0.5\textwidth]{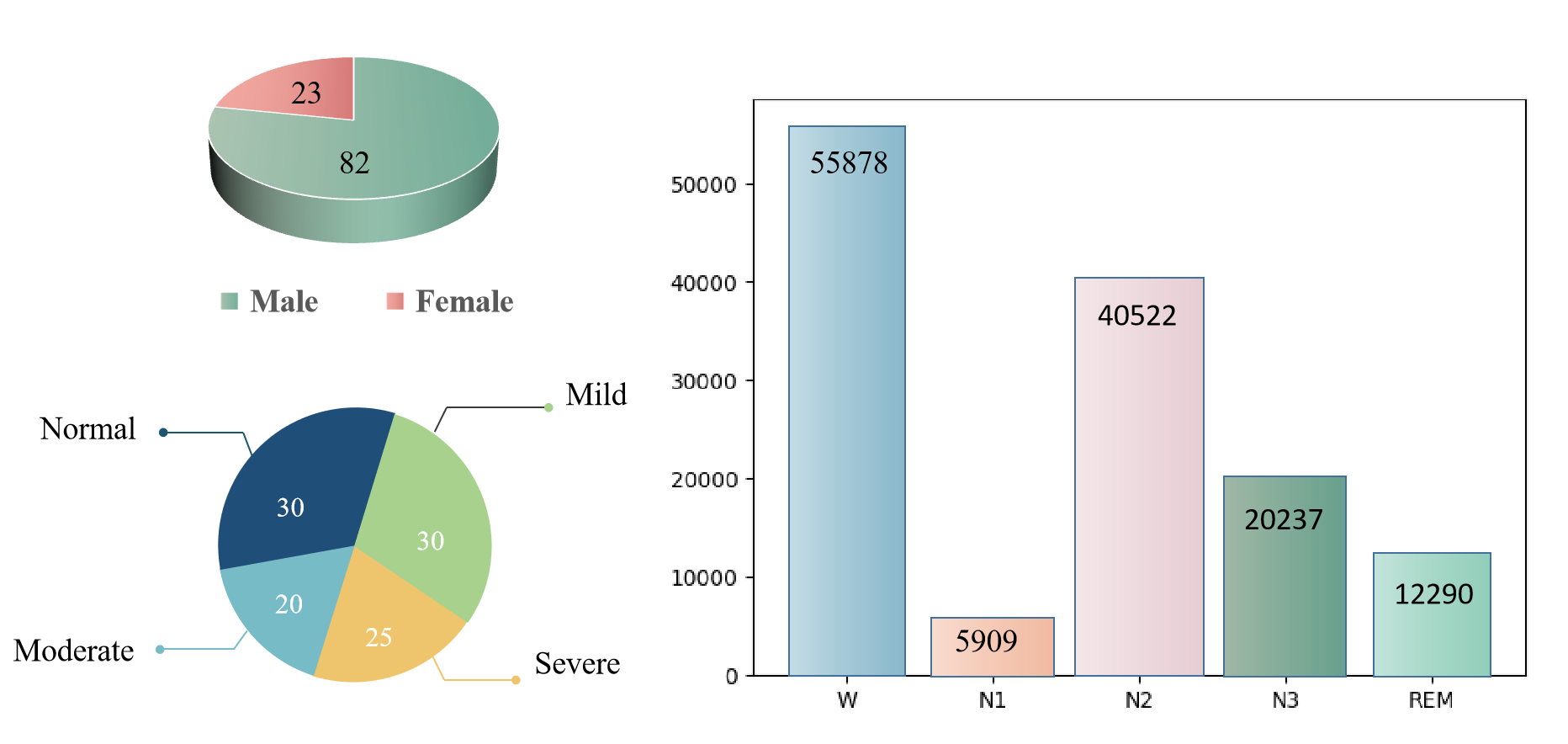}
  \caption{Statistical information on our dataset. The top left subfigure shows the distribution of male and female cases in the dataset; The bottom left subfigure shows the distribution of each AHI-related group case in the dataset; The right-hand subfigure shows the distribution of each sleep stage in the dataset.}
  \label{fig:dataset}
\end{figure}
\noindent\textbf{The time duration statistics.}
As shown in the upper left subfigure in Fig. \ref{fig:dataset}, we collected PSG signals and the synchronous infrared videos in our dataset, including the monitoring of 105 subjects' (82 Males and 23 Females) whole night data, consisting of 154,573 data clips. Each clip is 30 seconds long and the whole dataset contains 1124 hours' data. The number of clips per subject ranges from 1080 to 1360.

\noindent\textbf{The Apnea-Hypopnea Index (AHI) distribution.} The medical community usually classifies AHI meaningful into four clinically significant groups(\textless 5, 5-15, 16-30, \textgreater 30). The above four groups correspond to normal, mild obstructive sleep apnea (OSA), moderate obstructive sleep apnea (OSA), severe obstructive sleep apnea (OSA), respectively. As shown in the bottom left subfigure of Figure \ref{fig:dataset}, out of the 105 patients in our dataset, 30 are normal, 30 are mild, 20 are moderate, and 25 are severe.

\noindent\textbf{Sleep stage distribution.} Adult's sleep cycle lasts about 90 to 100 minutes, alternating about four to five times in a night. As shown in the right subfigure of the Fig. \ref{fig:dataset}, there are five stages, where $W$ represents wake period, $N1$ denotes the sleepy phase, which lasts about five minutes that describes the period between awake and falling asleep. $N2$ represents the period of light sleep, and with feeling of falling or weightlessness during sleep, as well as sudden body twitching. $N3$ is a deep sleep phase, in which brain wave activity drops to 1-2 seconds, and the respiration and heart rate reach the lowest. Deep sleep period accounts for about 20$\%$ of the sleep time per night. $REM$ is a period of rapid eye movement, in which the eyes begin to move rapidly and the blood pressure, heart rate and respiration rate are more active than in the Non-REM stage. As shown in the right subfigure in Figure \ref{fig:dataset}, in our dataset, there are 55,878, 5909, 40,522, 20,237 and 12,290 clips in $W$, $N1$, $N2$, $N3$ and $REM$ stages respectively.

\begin{figure*}
  \includegraphics[width=\textwidth]{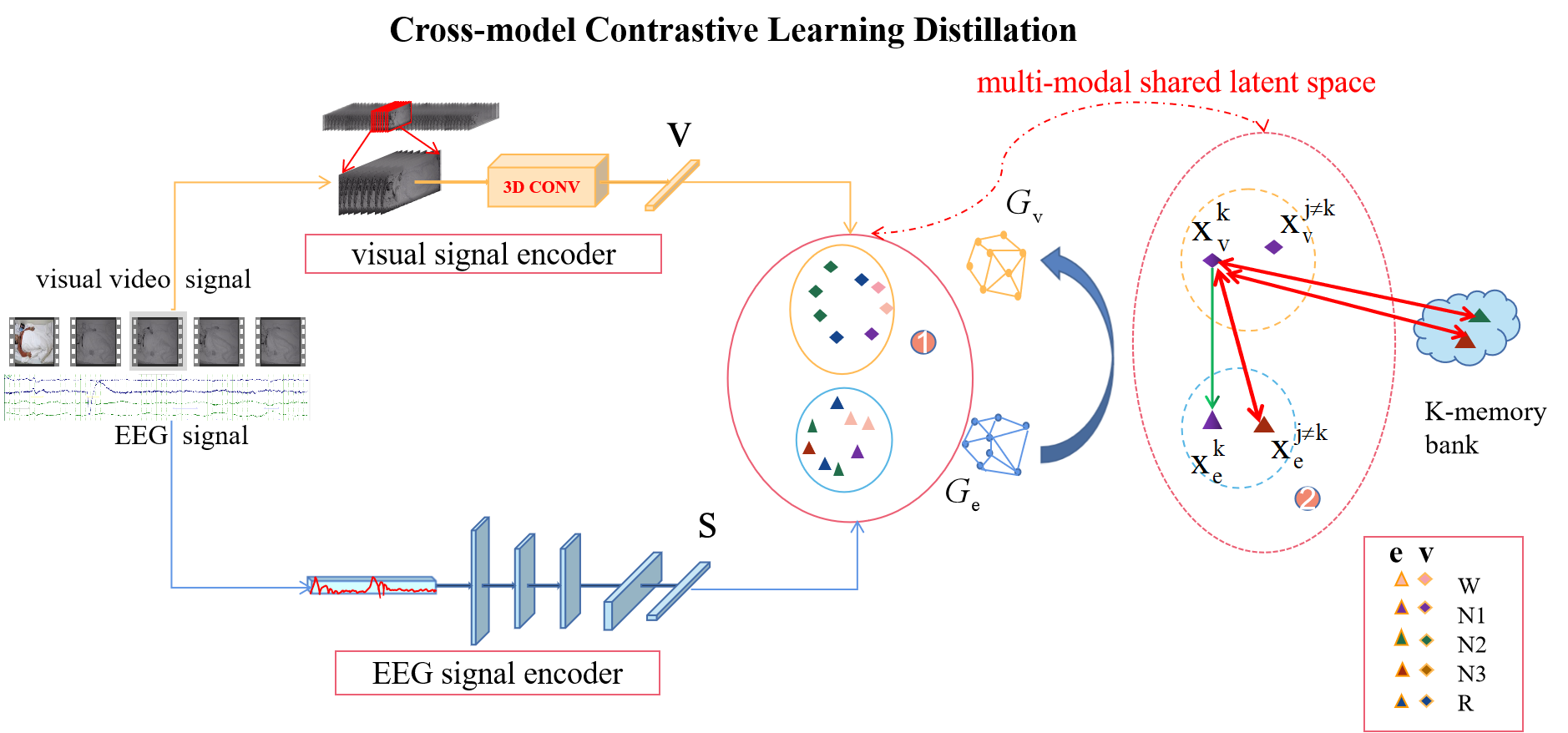}
  \caption{ Overall network architecture of cross-modal distillation.}
  \label{fig:architecture}
\end{figure*}

\section{Method}

\subsection{Overall Framework}\label{sec:4.1}
The overall architecture is shown in Fig. \ref{fig:architecture}, where we are given the $i^{th}$ synchronized EEG signal $s_{i}$, the ground-truth sleep stage annotation $y_{i}$ and the Infra-red video $v_{i}$ collected from real-world subjects in the hospital (see more analysis and statistics of the dataset in section \ref{sec:3.2}). Given the video encoder $E_v$ and EEG signal encoder $E_s$, the conventional cross-entropy loss $L_{CE}$ is utilized to train $E_v$, $E_s$ and the EEG and Infra-red classifiers $C_s$ and $C_v$ respectively as the following Eq.\eqref{eq:L_CE_coarse}: 
\begin{equation}
\label{eq:L_CE_coarse}
    \min_{E_v,E_s,C_v,C_s} \sum_{i} L_{CE}(C_v(E_v(v_{i})),y_{i}) + L_{CE}(C_s(E_s(s_{i})),y_{i}).
\end{equation}

The aim of our proposed method is not only to train an Infra-red video classification network, but to distill the cross-modal knowledge and discriminative structural information from the clinically-recognized ``gold-standard'' EEG feature $E_s(s)$ to the portable and point-of-care visual infra-red video feature $E_v(v)$. If it is achieved, we are able to classify the sleep stage using the infra-red video features only at home. Our intuition is that the infra-red video can capture the visual appearance patterns of the sleeping subjects but EEG signal cannot. However, the EEG signal can capture the intrinsic features of the brain electric signal that are clinical relevant but the videos cannot. Therefore, we argue that there exists a huge semantic gap between infra-red video feature distribution $P(E_v(v_{i}))$ and the EEG signal one $P(E_s(s_{i}))$ and their features complement each other in the sleep stage classification. 

\textbf{\textit{To distill the discriminative knowledge from $P(E_s(s_{i}))$  to $P(E_v(v_{i}))$, the first challenge is to tackle the cross-modal semantic gaps.}} We propose a structure-aware cross-modal distillation module, consisted of two graphical neural networks $G_v$ and $G_s$ to model the unimodal inter-sample relationships $G_v(E_v(v))$ and $G_s(E_s(s))$ respectively. Then we propose to reduce the cross-modal gaps of inter-sample relationships by reducing their structural distance loss $L_D$ based on a metric $D$. The optimization is shown in the following Eq.\eqref{eq:L_D_coarse} and more details are described in Section \ref{sec:4.2}.
\begin{equation}
\label{eq:L_D_coarse}
    \min_{E_v,E_s,G_v,G_s} \sum_{v,s} L_D(G_v(E_v(v)),G_s(E_s(s))).
\end{equation} 

Reducing the structural cross-modal gap does not align the fine-grained semantic concepts between EEG and visual embedding space. \textbf{\textit{To distill the discriminative and fine-grained knowledge from $P(E_s(s_{i}))$ to $P(E_v(v_{i}))$}}, we propose a cross-modal contrastive distillation framework utilizing two hard-negative memory selectors that stores the $K$-hardest negative samples based on the EEG and the video anchor samples. Specifically, our cross-modal contrastive distillation framework trains the encoders based on the contrastive learning loss $L_C(v_i,s_i,v_j,s_h)$ in Eq.\eqref{eq:L_C_coarse}, where $v_i,s_i$ are the $i^{th}$ cross-modal sample as the anchor points, while the $v_j,s_h$ are the negative pairs optimally selected based on our novel $K-$hardest negative sample selection module. The selection module consists of two memory queues designed for two modalities respectively. In our cross-modal contrastive distillation, the positive pairs are the synchronized features $<E_v(v_i),E_s(s_i))>$ but our selection module leveraged two negative pairs $v_j$ and $s_h$, that are selected optimally from EEG and the visual memory queue $Q_s$ and $Q_v$ respectively. The selection criterias are as follows: for a cross-modal target anchor pair $v_i,s_i$, the hard negative pair are selected as $v_j$ and $s_h$ respectively, where $j = argmax({E_s(s_i)}^TE_v(v_j)), for \; \forall v_j \in Q_v$ and $h = argmax({E_v(v_i)}^TE_s(s_h)), for \; \forall s_h \in Q_s$. This procedure maintains and aligns the fine-grained cross-modal semantic embedding. More details are shown in Section \ref{sec:4.3}. 
\begin{equation}
 \label{eq:L_C_coarse}
    \min_{E_v,E_s} \sum_{v,s} L_C(E_v(v_i),E_v(v_j),E_s(s_i),E_s(s_h)).
\end{equation}

Besides the feature space semantic alignment, we also adopt the conventional class prediction space distillation as shown in the following Eq.\eqref{eq:L_JSD_coarse}. We utilized and reduced the well-known Jenson-Shannon-Divergence (JSD) between the visual prediction distribution $P(C_v(E_v(v)))$ and the EEG one $P(C_s(E_s(s)))$. The overall optimization is shown in Eq.\eqref{eq:L_overall} where $\lambda_1$,$\lambda_2$ and $\lambda_3$ are hyper-parameters. 

\begin{equation}
 \label{eq:L_JSD_coarse}
    \min_{E_v,E_s,C_v,C_s} JSD(P(C_v(E_v(v))),P(C_s(E_s(s)))).
\end{equation}

\begin{equation}
 \label{eq:L_overall}
    \min L_{CE} + \lambda_1 L_{D} + \lambda_2 L_{C} + \lambda_3 JSD.
\end{equation}

\subsection{Structure-aware coarse-grained semantic alignment} \label{sec:4.2}

To distill the knowledge from clinical signal to IR videos, the two synchronized data in our $S^3VE$ are less correlated and with a large semantic gap than those in the conventional audio-video classification benchmark, e.g. UCF51 \cite{soomro2012dataset}. Therefore, we argue that directly pulling the cross-modal features of the same \textit{instance} and pushing the features of different \textit{instances} may lead to inferior distillation performance and the collapsed joint embedding space. 

To tackle the previous challenge, \textit{we propose to align the relationship structure formed by multiple cross-modal clip features instead of from the individual instance}. Our intuition is that in spite of the instance-level semantic gap, the uni-modal structure relationships should be similar among multiple instances as a group. In addition, the cross-modal data exhibit unique temporal characteristics and a strong temporal correlation among the features in a mini-batch is observed. Formulating both points as a constraint, we propose to build a graph among multiple instances in each unimodal data and regularize the consistency of two unimodal relationship structures.

More specifically, given the $i^{th}$ clip of an input mini-batch, $E_s(x_{i})$
and $E_v(x_{i})$ denotes the output features of two modalities' graph level encoders. Below, we take the video and EEG signal $v_i$, $e_i$ as an example, Each graph $G=(V,E)$ is represented as sets of nodes $V$ and edges $E$, 

\begin{equation}
 \label{eq:1}
    h_{{v_i}}^{(0)} = {\rm{ML}}{{\rm{P}}_{{\rm{node}}}}\left( {{x_{{v_i}}}} \right)\\
\end{equation}       
  
\begin{equation}
 \label{eq:1}
    {e_{ij}} = {\rm{ML}}{{\rm{P}}_{{\rm{edge}}}}\left( {{x_{v_{ij}}}} \right)
\end{equation}       
The output of $E_v(x_{i})$ : $ h_{{v_i}}^{(0)}$ is a  64-dimensional vector,through an MLP, the hidden node vector $ h_{{v_i}}^{(0)}$ 's dimension is 128, and the edges vector $e_{ij}$'s dimension is 64 .
Nodes are transfer information in the propagation layers. After the $t-th$ pass, the propagation layer maps a node representations $ h_{i}^{(t)}$ to new node representations  $h_{i}^{(t+1)}$, here node's dimension have not change. $F_{m} $ is an MLP work on the concatenated inputs of $h_{i}^{(t)}$, $ h_{j}^{(t)}$ and  $e_{i j}$, here the output vector dimension here is 512-dim.
then we utilize weighted summation based on graph attention mechanism $A$ \cite{velivckovic2017graph} to update node features. Through multiple layers of propagation, the representation for each node will accumulate information in its local neighborhood.

\begin{equation}
     m_{j \rightarrow i}=F_{m}\left(h_{i}^{(t)}, h_{j}^{(t)}, e_{i j}\right)
\end{equation}
Here $F_{m}$ is an MLP on the concatenated inputs including neighbor node features and edge features.
\begin{equation}
h_{i}^{(t+1)}=A\left(h_{i}^{(t)} ,\sum_{j \in \delta_{i}} m_{j \rightarrow i}\right)
\end{equation}
After the $t-th$ layer propagation, a graph level representation $O_{\left(G_{v})\right.}$ can form through a $READOUT$ function and a graph level $ MLP_{G}$ , and here we choose a fully connected layer as the READOUT function. $O_{\left(G_{v})\right.}$ is a 1024-dim vector.
 \begin{equation}
    O_{(G_{v})}= MLP_{G}(READOUT\{h_{i}^{(t)}: X_{v_{i}} \} \in V)
 \end{equation}
we use Euclidean similarity to compute the structural distance loss $L_D$ between $G_v$ and $G_a$ :
\begin{equation}
\begin{aligned}
  L_D=E_{(G_e,G_v)}(max,r-d(G_v,G_a))\\
  d(G_V,G_a)={||O_{(G_v)}-O_{(G_v)}||}^2
\end{aligned}
\end{equation}
where $\gamma>0$ is a margin parameter.

\begin{figure}[t]
  \includegraphics[width=0.48\textwidth]{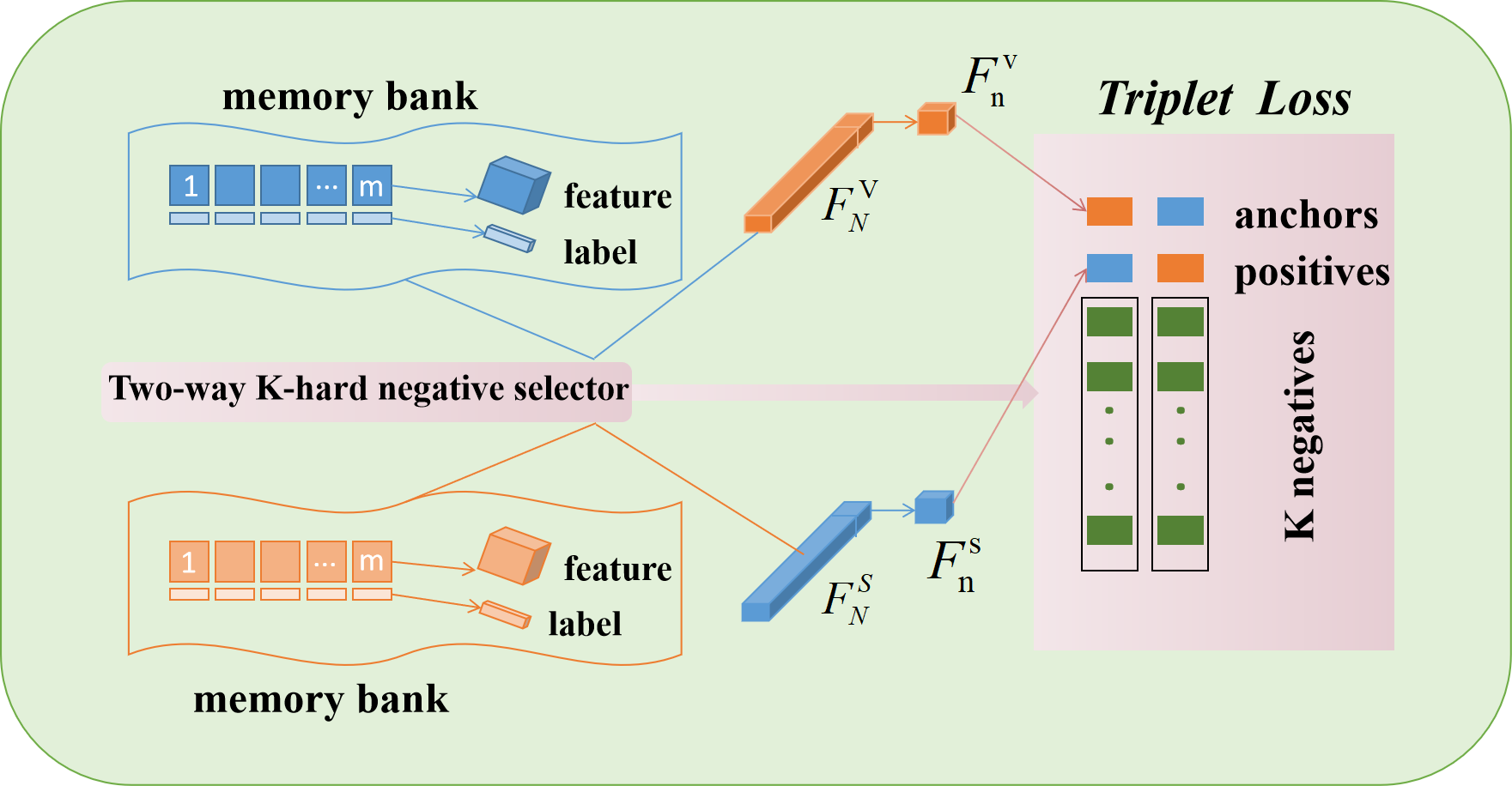}
  \caption{ Two K-hard negative memory bank structure diagram, and a momentum updated Encoder to generate momentum updated embeddings stored in a large memory bank.}
  \label{fig:architectureKkkkk}
\end{figure}

\subsection{Hard-negative fine-grained cross-modal contrastive learning} \label{sec:4.3}


As mentioned in the introduction section, the IR video appearance features are less discriminative than the EEG signal, especially since the inter-class differences are locally subtle in the video. To distil the discriminative EEG feature to the ambiguous visual ones, usage of the contrastive learning strategy--pulling positive pairs and pushing away the negative ones seems a straightforward choice. However, conventional cross-modal contrastive learning is limited in our scenario because the negative pairs are normally selected from the cross-modal features of different instances, which have large semantic gaps. Therefore, the normal negative pairs are not ``hard'' enough, which cannot reflect and reveal the fine-grained contrast required in our setup. Furthermore, the current contrastive learning method usually obtains negative samples in the same batch, and the number of negatives directly affects the contrastive loss. For our task, since the distribution of clips within a batch is usually flat, the number of negative samples is insufficient, and it is possible to take negative samples within one class to indirectly affect the classification performance. 

To tackle the challenges, \textit{we propose two $K$-hard negative memory bank Figure \ref{fig:architectureKkkkk} for the EEG and visual modality respectively, selecting the negative sample set in the online manner. In addition, we propose to use the symmetric contrastive cross-modal distillation to reduce the cross-modal semantic gap.}

Specifically, each $K$-hard negative memory bank stores the first $K$ hardest negative pairs for each class categories, preserving the global hard negative sample features in the dataset level. This helps ensure that the contrast in the optimization are fine-grained. In addition, to shrinkage the cross-modal semantic gap, we proposed the two-way symmetrical contrastive learning loss as shown in the following equation\eqref{two-k} :
\begin{equation}
\begin{aligned}
  L_{D_N}=\sum_{n=1}^{N}\{[\alpha-S(F_n^v,F_n^s)+S(F_n^v,F_i^s)]_+\\
+[\alpha-S(F_n^v,F_n^s)+S(F_j^v,F_n^s)]_+\}
\end{aligned}
\label{two-k}
\end{equation}
where $\alpha$ is the margin we set 0.2 here and can be tuned using the validation sets. $S(\cdot)$ is the similarity function in the feature space. $F_n^v$, $E_n^s$ is the output feature from encoder $E_v$ and $E_s$ , which are consistent with the description in \ref{sec:4.1}. $i$ and $j$ are the index for the hard negatives , $n$ is the anchor index for distillation and $N$ denotes the batch size.

Specifically, we used two $K$-hard negative memory bank to enlarge the hard negative sample set, the memory bank $ M_B$ stores the feature representation of historical samples with a size of $m$. We rewrite the formula as:
\begin{equation}
\begin{aligned}
  L_{D_N}^K &=\sum_{k=1}^{K}\sum_{n=1}^{N}\{[\alpha-S(F_n^v,F_n^s)+S(F_n^v,F_k^s)]_+ \\&+[\alpha-S(F_n^v,F_n^s)+S(F_k^v,F_n^s)]_+\}
\end{aligned}
\end{equation}
During the network training, we dynamically update the memory bank by discarding the oldest items and feeding the new batch of embedded features, where the memory bank acts as a queue. And we also keep the class label along with representation in the memory bank to filter the negatives.The update method of memory bank we use is  Momentum updated Encoder (MuEncoder) \cite{wang2020learning}. The memory bank structure is shown in the figure \ref{fig:architectureKkkkk}.

\section{Experiment}
\subsection{Dataset Splits}

Because some samples have individual reasons (short sleep time, difficulty falling asleep, strange posture, etc.), we reduced the samples by 3 to 102 subjects.

According to the sleep apnea indexes (no apnea, mild apnea, moderate apnea, severe apnea), samples at each apnea stage are subdivided, and 80\% of each apnea stage is taken as the training set (Among them, 20\% were chosen as the validation set), the last 20\% of each apnea stage is taken as the testing set. Twenty people are used for the test to ensure that the training and testing are equally distributed. Therefore, the clip number of the train set is 102,519, the clip number of the validation set is 20,503, and the clip number of the test set is 31,551.

\subsection{Implementation Details and Metrics}
\label{details}
To be fair and comprehensive, we use R(2+1)D-18 \cite{tran2018closer} and R3D-18 \cite{hara2017learning} as the video student network, which is pretrained in Kinetics-700 \cite{carreira2019short} and Moments in Time \cite{monfort2019moments} and fine-tuned in our dataset ($S^3VE$). We use AttnSleep \cite{eldele2021attention} as the EEG teacher network, which is only trained on our dataset, and the model weights of the teacher network are kept frozen during training. The video network baseline is trained by SGD with a learning rate of 0.001, and a weight decay of 0.0005. When training, our batch size is set to 16; we trade off computational efficiency and set the size of the memory bank as 256, the number of negatives $K$ as 64. The hyperparameters $\lambda_1$, $\lambda_2$ and $\lambda_3$ are set to 0.5, 1 and 1. The dimension of the latent feature space is 512. We do not add projections on the network, but linear projections can be added to map all embeddings to the same dimension. The video clips are cropped to 320×240 and each clip contains about 750 frames. During the testing phase, just the IR sleep video was used for classification.

We adopted the following three evaluation metrics to measure the performance of the sleep classification: the accuracy (ACC), macro-averaged F1-score (MF1) and Cohen Kappa ($\kappa$) \cite{cohen1960coefficient}.They are calculated as follows:
\begin{equation}
\begin{aligned}
    ACC=\frac{ \sum_{c\in{C^S}}TP_c}{N}
\end{aligned}
\label{ACC}
\end{equation}
\begin{equation}
\begin{aligned}
    MF1=\frac{ \sum_{c\in{C^S}}F1_c}{5}
\end{aligned}
\label{MF1}
\end{equation}
where $TP_c$ and $F1_c$ are the true positive and per-class F1 score of class $c \in$ $C^S$, respectively, and $N$ is the total number of test
samples. $\kappa$ is a statistical measure of the interrater agreement (IRA) level calculated as:
\begin{equation}
\begin{aligned}
    \kappa=\frac{ \sum_{c\in{C^S}}{p_{cc}} -\sum_{c\in{C^S}}{p_{c+}}{p_{+c}}}{1 -\sum_{c\in{C^S}}{p_{c+}}{p_{+c}}}=\frac{p_a-p_e}{1-p_e}
\end{aligned}
\label{K}
\end{equation}
where $p_cc$ represents the percentage of epochs classified as category c by the network and the annotated label simultaneously,and $p_c+$ and $p_+c$ represent the percentages of epochs classified as category $c$ by the network and annotated label, respectively.

\begin{table*}[t]
\small
\caption{Comparison Results (\%) among SACD and State-of-art Models. The Best Values on $S^3VE$ Dataset Are Highlighted in Bold.\\The number following $\pm$ represents the standard deviation of multiple experiments.}
\centering{
\resizebox{1\linewidth}{!}{
\begin{tabular}{l|c|lclclcl|lclclclclc}
\hline
                                  & \multicolumn{1}{l|}{} & \multicolumn{7}{c|}{Overall Metrics}                                     & \multicolumn{10}{c}{Per-class Accuracy}                                                                               \\
EEG Baseline                      & Distillation Method   &  & Accuracy           &  & MF1                &  & $\kappa$           &  &  & W                  &  & N1                 &  & N2                 &  & N3                 &  & REM                \\ \hline
\multicolumn{1}{c|}{-}            & Baseline R(2+1)D      &  & 52.5               &  & 46.6               &  & 0.47               &  &  & 64.7               &  & 41.3               &  & 48.4               &  & 47.8               &  & 57.3               \\
\multicolumn{1}{c|}{-}            & Baseline R3D          &  & 52.8               &  & 46.9               &  & 0.37               &  &  & 64.7               &  & 41.4               &  & 49.0               &  & 48.0               &  & 58.0               \\ \cline{1-1}
\multicolumn{1}{c|}{}             & Fitnet                &  & 56.1               &  & 49.9               &  & 0.49               &  &  & 68.5               &  & 47.2               &  & 55.9               &  & 53.8               &  & 61.8               \\
                                  & PKT                   &  & 58.1               &  & 51.2               &  & 0.53               &  &  & 71.2               &  & 47.8               &  & 56.3               &  & 54.0               &  & 62.9               \\
                                  & CRD                   &  & 62.3               &  & 55.8               &  & 0.56               &  &  & 73.8               &  & 49.8               &  & 58.8               &  & 57.1               &  & 66.4               \\
\multicolumn{1}{c|}{Attnsleep}    & IFD                   &  & 61.1               &  & 54.0               &  & 0.57               &  &  & 73.0               &  & 49.9               &  & 58.8               &  & 57.1               &  & 65.1               \\
                                  & CMC                   &  & 58.9               &  & 53.1               &  & 0.53               &  &  & 71.5               &  & 46.2               &  & 56.9               &  & 55.0               &  & 63.2               \\
                                  & CCL                   &  & 62.3               &  & 57.0               &  & 0.56               &  &  & 74.4               &  & 50.1               &  & 58.7               &  & 53.8               &  & 64.7               \\
                                  & SACD(ours)            &  & \textbf{64.4$\pm$0.45}          &  & \textbf{58.9$\pm$0.40} &  & \textbf{0.60$\pm$0.38} &  &  & \textbf{75.6$\pm$0.56} &  & \textbf{51.0 $\pm$0.36} &  & \textbf{60.2$\pm$0.42} &  & \textbf{59.3$\pm$0.45} &  & \textbf{67.8+0.45} \\ \hline
                                  & Fitnet                &  & 54.6               &  & 47.8               &  & 0.46               &  &  & 67.0               &  & 45.2               &  & 51.9               &  & 49.8               &  & 59.5               \\
                                  & PKT                   &  & 55.7               &  & 48.6               &  & 0.49               &  &  & 69.6               &  & 46.2               &  & 52.3               &  & 50.1               &  & 59.8               \\
                                  & CRD                   &  & 58.3               &  & 51.9               &  & 0.50               &  &  & 70.2               &  & 47.8               &  & 55.0               &  & 53.2               &  & 62.4               \\
\multicolumn{1}{c|}{DeepSleepNet} & IFD                   &  & 57.2               &  & 51.3               &  & 0.49               &  &  & 69.8               &  & 46.8               &  & 54.0               &  & 53.4               &  & 61.2               \\
                                  & CMC                   &  & 55.5               &  & 50.8               &  & 0.48               &  &  & 68.5               &  & 44.8               &  & 53.0               &  & 51.0               &  & 59.5               \\
                                  & CCL                   &  & 58.7               &  & 52.3               &  & 0.52               &  &  & 71.0               &  & 47.7               &  & 54.7               &  & 50.8               &  & 61.0               \\
                                  & SACD(ours)            &  & \textbf{60.7$\pm$0.52} &  & \textbf{54.6$\pm$0.39} &  & \textbf{0.54$\pm$0.45} &  &  & \textbf{72.2+0.60} &  & \textbf{48.4$\pm$0.36} &  & \textbf{56.2$\pm$0.38} &  & \textbf{55.3$\pm$0.42} &  & \textbf{63.8$\pm$0.45} \\ \hline
                                  & Fitnet                &  & 54.9               &  & 48.9               &  & 0.48               &  &  & 67.7               &  & 45.5               &  & 52.2               &  & 50.1               &  & 60.0               \\
                                  & PKT                   &  & 57.5               &  & 51.0               &  & 0.50               &  &  & 51.4               &  & 48.0               &  & 54.1               &  & 51.7               &  & 61.6               \\
                                  & CRD                   &  & 60.5               &  & 53.9               &  & 0.53               &  &  & 72.6               &  & 48.7               &  & 57.0               &  & 55.1               &  & 64.6               \\
\multicolumn{1}{c|}{Modified-SEN} & IFD                   &  & 59.0               &  & 53.0               &  & 0.51               &  &  & 71.7               &  & 48.6               &  & 55.6               &  & 54.8               &  & 63.0               \\
                                  & CMC                   &  & 57.0               &  & 51.1               &  & 0.51               &  &  & 70.0               &  & 46.6               &  & 54.5               &  & 52.4               &  & 61.2               \\
                                  & CCL                   &  & 60.6               &  & 54.2               &  & 0.53               &  &  & 72.8               &  & \textbf{49.6}      &  & 56.5               &  & 52.7               &  & 62.8               \\
                                  & SACD(ours)            &  & \textbf{62.6$\pm$0.45} &  & \textbf{55.6$\pm$0.48} &  & \textbf{0.56$\pm$0.38} &  &  & \textbf{73.8$\pm$0.52} &  & 49.5$\pm$0.45          &  & \textbf{58.5$\pm$0.55} &  & \textbf{57.3$\pm$0.48} &  & \textbf{66.1$\pm$0.36} \\ \hline
\end{tabular}}
}
\label{tab: mainnew}
\end{table*}

\begin{table*}[t]
\small
\caption{Video classification on the Public dataset UCF51. Metric:Top1 accuracy (\%). Knowledge is transferred from audio modality to improve the video recognition model.}
\centering
{
\begin{tabular}{c|lllllllclllllllc}
\hline
\multicolumn{1}{l|}{} & \multicolumn{16}{c}{UCF51}                                                                                                                                                                                                  \\
Methods               &  & Baseline R(2+1)D         &  & Fitnet                   &  & \multicolumn{1}{c}{PKT} &  & RKD  &  & \multicolumn{1}{c}{CRD} & \multicolumn{1}{c}{} & \multicolumn{1}{c}{CMC} &  & \multicolumn{1}{c}{CCL} &  & SACD(ours) \\ \hline
Audio to Video        &  & \multicolumn{1}{c}{57.5} &  & \multicolumn{1}{c}{48.4} &  & 53.2                    &  & 53.0 &  & 60.3                    &                      & 59.2                    &  & 64.9                    &  & 66.0       \\ \hline
\end{tabular}

}
\label{tab:ucf51}
\end{table*}

\subsection{Results and Comparisons}
\label{results}
We compare our SACD with the only infrared video baseline models without any distillation, and six state-of-the-art other distillation methods \textbf{(Fitnet  \cite{romero2014fitnets}, PKT \cite{passalis2018learning}, CRD \cite{tian2019contrastive}, IFD \cite{passalis2020heterogeneous}, CMC \cite{tian2020contrastive}, CCL\cite{chen2021distilling})}. For the fairness and comprehensiveness of the experiments, we train each model on the same experimental setup but replace the distillation objective based on their open-source implementation.

As shown in Table \ref{tab: mainnew}, our method achieves state-of-art in the accuracy (ACC), macro-averaged F1-score (MF1), Cohen Kappa ($\kappa$) and per-class accuracy. By comparison, we found that CRD \cite{tian2019contrastive} and CCL \cite{chen2021distilling} are the two most powerful opponents. Both methods apply contrastive learning, which indirectly shows that contrastive learning can significantly improve the cross-modal tasks. However, we are using a new contrastive learning method which has enabled us to achieve an overall lead in all evaluation metrics. Specifically, our method SACD is 11.9\% (52.5\%-64.4\%) higher than the baseline in terms of accuracy, and there is also a 12.3\% (46.6\%-58.9\%) and 0.13 (0.47-0.60) increase in both MF1 and $\kappa$, respectively. Compared to our next closest rival CCL, we are also 2.1\% (62.3\%-64.4\%), 1.9\% (57.0\%-58.9\%) and 0.04 (0.56-0.60) higher in ACC, MF1 and $\kappa$, respectively; and our method is also higher than the CCL in all five separate sleep stages, as shown in the TABLE \ref{tab: mainnew}. The above results show that our method SACD has superior performance on our dataset $S^3VE$ and outperforms other SOTA methods under various evaluation metrics.

To demonstrate that our cross-modal distillation method works not only under a single EEG modality baseline (AttnSleep), we also performed experiments similar to the previous section under more SOTA EEG baselines. Therefore, we chose DeepSleepNet \cite{supratak2017deepsleepnet} and a single-channel version of SEN-DAL \cite{jia2022multi} (no EOG signal is input, only a single-channel EEG signal is input, and the two output heads of Label Prediction and Domain Classification are still retained) as the EEG modality baseline, and the experimental results are shown in TABLE \ref{tab: mainnew}. When DeepSleepNet is selected as the EEG modality's baseline, our SACD exceeds other cross-modal distillation methods, including CRD and CCL, in the $S^3VE$ dataset. Specifically, compared to the IR video modality baseline, SACD outperforms ACC, MF1 and $\kappa$ by 6.1\% (54.6\%-60.7\%), 6.8\% (47.8\%-54.6\%) and 0.08 (0.46-0.54), respectively. In addition to this, SACD is superior to existing SOTA methods in per-class accuracy. When a single-channel version of SEN-DAL (Modified-SEN) is chosen as the baseline for our EEG modality, our SACD outperforms the baseline by 7.7\% (54.9\%-62.6\%), 6.7\% (48.9\%-55.6\%) and 0.08 (0.48-0.56) for ACC, MF1 and kappa, respectively. Furthermore, compared to our most competitive rival CCL, our SACD improves by 2\% (60.6\%-62.6\%), 1.4\% (54.2\%-55.6\%) and 0.03 (0.53-0.56) for ACC, MF1 and $\kappa$, respectively. In terms of per-class accuracy, our methodology is significantly better than the current one, except for the $N1$ stage, where our SACD is slightly lower than the CCL.

To illustrate that our proposed distillation method is not only applicable to our $S^3VE$ dataset, we also conduct experiments on the UCF51 \cite{soomro2012dataset} public dataset. UCF51 is a subset of UCF101 that contains audio in videos, including 6,845 videos from 51 action classes. We use the public split 1 for evaluation. We choose the pre-trained audio encoder 1D-CNN14 \cite{kong2020panns} trained on  AudioSet \cite{gemmeke2017audio} as the teacher and freeze its parameters. The experimental results are shown in  TABLE \ref{tab:ucf51} and our SACD also obtains the state-of-the-art consistently and improves over the prior methods. Specifically, in terms of ACC, our SACD improved by 8.5\% (57.5\%-66.0\%) over the baseline method and by 1.1\% (64.9\%-66.0\%) over CCL. The results show that our method can also achieve superior distillation results on other datasets.

In general, the above-mentioned mainstream distillation methods we compare include single-modal distillation and cross-modal distillation, and the latter three methods also apply contrastive learning. Compared with them, we first introduce structural similarity across modalities, aiming to improve the ability to deal with weak inter-class gaps while narrowing the semantic gap. Secondly, we apply two $K$-hard negative samples to train highly transferable sample visual representations. Experimental results demonstrate that our method significantly improves cross-modal distillation on the $S^3VE$ dataset. It also indicates that it can do sleep stage classification with reasonable accuracy from IR sleep videos alone.

\subsection{Evaluation of sleep stage using both the EEG and IR modality}
Our starting point is to distill the knowledge of the EEG modality to the IR video, and we also verified our conjecture using experiments In section \ref{results}. This section will analyze the complementary information in EEG and IR modalities. In other words, is it beneficial to provide better predictions if we combine information from these two modalities? As EEG and IR data are naturally synchronized in the data collection, the simplest and most effective fusion method is the late-fusion on the features that can fully consider the interactions and correlations between each modality. We use AttnSleep \cite{eldele2021attention} as the EEG teacher network trained on the $C3$ (Channel 3) data in the $S^3VE$ training set for 58 epochs. The five-class sleep stage classification using EEG data achieves the accuracy of 78.8\% on the validation set. For the IR baseline, we chose R3D-18, trained for 80 epochs on our dataset $S^3VE$. As shown in the second row of TABLE \ref{tab: mainnew}, the sleep stage classification accuracy for IR video is 52.8\%. We first save the weights of the two networks, then perform the inference operation on the 80 subjects of the training set and save the encoded features of the EEG encoder and IR video encoder of each clip. The two model output feature dimensions must be consistent, and we perform two experiments using 512 and 3000 dimension features as the output, respectively. The experimental results show little difference between the results of the two choices.

\begin{table}[]
 \small
 \centering
\caption{Sleep stage classification for IR+EEG. Metric:Top1 accuracy (\%).
   \textit{ (In Experiment A, We use a single fully connected layer as the prediction head; In Experiment B, We use three fully connected layers as the prediction head. Experiment C is a version of the released gradients from Experiment A; Experiment D is a version of the released gradients from Experiment B. In experiment E, we use our SACD to distill IR video information (frozen gradients) onto the EEG baseline. In experiment F, we use our SACD to distill IR video information (released  gradients) onto the EEG baseline. In experiment G, we use CCL to distill IR video information (released gradients) onto the EEG baseline)}}
\setlength\tabcolsep{0.55pt}
\resizebox{0.8\linewidth}{!}{
\begin{tabular}{cl|lcl|llllllllllllll}
\hline
\multicolumn{1}{l}{}                                                     &  &  & \multicolumn{1}{l}{}         &  & \multicolumn{14}{c}{Per-class Accuracy}                                                                                                                                                                                \\
Methods                                                                  &  &  & \multicolumn{1}{l}{Accuracy} &  &  & \multicolumn{1}{c}{W} &                       &  & \multicolumn{1}{c}{N1} &                       &  & \multicolumn{1}{c}{N2} &                       &  & \multicolumn{1}{c}{N3} &                       &  & REM  \\ \hline
\begin{tabular}[c]{@{}c@{}}Baseline R3D\\  (only IR video)\end{tabular}  &  &  & 52.8                         &  &  & 64.7                  & \multicolumn{1}{l|}{} &  & 41.3                   & \multicolumn{1}{l|}{} &  & 48.4                   & \multicolumn{1}{l|}{} &  & 47.8                   & \multicolumn{1}{l|}{} &  & 57.3 \\ \hline
\begin{tabular}[c]{@{}c@{}}Baseline Attnsleep\\  (only EEG)\end{tabular} &  &  & 78.8                         &  &  & 92.8                  & \multicolumn{1}{l|}{} &  & 54.1                   & \multicolumn{1}{l|}{} &  & 74.2                   & \multicolumn{1}{l|}{} &  & 78.7                   & \multicolumn{1}{l|}{} &  & 70.4 \\ \hline
Experiment A                                                             &  &  & 79.3                         &  &  & 93.4                  & \multicolumn{1}{l|}{} &  & 54.8                   & \multicolumn{1}{l|}{} &  & 74.5                   & \multicolumn{1}{l|}{} &  & 78.7                   & \multicolumn{1}{l|}{} &  & 70.8 \\ \hline
Experiment B                                                             &  &  & 80.4                         &  &  & 94.0                  & \multicolumn{1}{l|}{} &  & 55.5                   & \multicolumn{1}{l|}{} &  & 75.3                   & \multicolumn{1}{l|}{} &  & 79.4                   & \multicolumn{1}{l|}{} &  & 71.8 \\ \hline
Experiment C                                                             &  &  & 84.4                         &  &  & 96.1                  & \multicolumn{1}{l|}{} &  & 59.2                   & \multicolumn{1}{l|}{} &  & 80.2                   & \multicolumn{1}{l|}{} &  & 84.3                   & \multicolumn{1}{l|}{} &  & 76.1 \\ \hline
Experiment D                                                             &  &  & 85.5                         &  &  & 96.9                  & \multicolumn{1}{l|}{} &  & 60.5                   & \multicolumn{1}{l|}{} &  & 81.8                   & \multicolumn{1}{l|}{} &  & 85.6                   & \multicolumn{1}{l|}{} &  & 77.4 \\ \hline
Experiment E                                                             &  &  & 80.6                         &  &  & 94.2                  & \multicolumn{1}{l|}{} &  & 55.5                   & \multicolumn{1}{l|}{} &  & 75.6                   & \multicolumn{1}{l|}{} &  & 79.8                   & \multicolumn{1}{l|}{} &  & 72.2 \\ \hline
Experiment F                                                             &  &  & 85.3                         &  &  & 96.7                  & \multicolumn{1}{l|}{} &  & 60.1                   & \multicolumn{1}{l|}{} &  & 81.5                   & \multicolumn{1}{l|}{} &  & 85.3                   & \multicolumn{1}{l|}{} &  & 77.4 \\ \hline
Experiment G                                                             &  &  & 83.2                         &  &  & 95.0                  & \multicolumn{1}{l|}{} &  & 58.7                   & \multicolumn{1}{l|}{} &  & 79.6                   & \multicolumn{1}{l|}{} &  & 82.0                   & \multicolumn{1}{l|}{} &  & 74.3 \\ \hline
\end{tabular}
}
\label{tab:ireeg}
\end{table}

As shown in TABLE \ref{tab:ireeg}, high accuracy (78.8\% for overall $ACC$, 92.83\% for stage $W$, 54.18\% for stage $N1$, 74.16\% for stage $N3$, 78.74 for stage $N3$, and 70.48\% for stage $REM$) can already be achieved using only the EEG baseline, because EEG is the ``gold-standard'' in sleep stage classification. In experiment $A$, we concatenate the two 512-dimensional tensors saved by the EEG and IR video encoder and use a fully-connected layer with an input dimension of 1024 and an output dimension of 5 as the prediction module. We train the network for 100 epochs, and the results are in TABLE \ref{tab:ireeg}. It is easy to observe that the accuracy improvement of various classes is not obvious. Specifically, 0.5\% (78.8\%-79.3\%) for overall accuracy; 0.6\% (92.8\%-93.4\%) for stage W; 0.6\% (54.2\%-54.8\%) for stage N1; 0.3\% (74.2\%-74.5\%) for stage N2; 0\% (78.7\%-78.7\%) for stage N3; 0.4\% (70.4\%-70.8\%) for stage REM.
We argue that the slight improvement in classification accuracy might be because the parameters of a single MLP classification module are not big enough to cover the dataset. Then we select more fully-connected layers for feature-level fusion. As shown in TABLE \ref{tab:ireeg}, in experiment $B$, we use three fully-connected layers as the prediction module. Their input and output dimensions are:
\begin{itemize}
 \item Input 1024 dimensions, output 256 dimensions.
\item Input 256 dimensions, output 64 dimensions.
\item Input 64 dimensions, output 5 dimensions.
\end{itemize}

We use the SGD to train this prediction module for 100 epochs and the results are significantly improved compared to Experiment $A$. Specifically, experiment $B$ obtains 1.3\% (78.8\%-80.1\%) improvement in overall accuracy; 1.3\% (92.8\%-94.1\%) for stage $W$; 1.4\% (54.1\%-55.5\%) for stage N1; 1.1\% (74.2\%-75.3\%) for stage $N2$; 1.1\% (78.7\%-79.8\%) for stage N3; 1.3\% (70.5\%-71.8\%) for stage $REM$. In both Experiments $A$ and $B$, we freeze the gradients of the encoders and train only the later linear projection. In contrast, in Experiments $C$ and $D$, we release the gradients of the encoders. The experimental results show that the accuracy of the sleep grading has a significant improvement of 5.6\% (78.8\%-84.4\%) and 6.7\% (78.8\%-85.5\%), respectively, over using only the EEG baseline AttnSleep.

To observe the performance of reverse distillation from the IR video to the EEG modality, we again design Experiment $E$ and Experiment $F$, which are the frozen gradients version and the released gradients version, respectively. It can be observed that the results of reverse distillation reached 80.6\% and 85.3\%, which is already extremely close to the fusion version of the experiment. In addition, we also perform a complementary experiment $G$ (the released gradients version of the CCL distillation method based on the reverse distillation from IR video to EEG modality), whose overall accuracy is 2.1\% (83.2\%-85.2\% ) lower than ours for the same setup.

From these results, we can draw the following conclusions:

\begin{itemize}
\item \textbf{Sleep stage classification accuracy improves by using the feature-level fusion of EEG and IR videos.}
\item \textbf{Most importantly, we show that complementary information exists between the two modalities. In other words, studying the cross-modal distillation of IR and EEG modalities makes sense.}
\item \textbf{If the cross-modal distillation method is good enough, the accuracy of distillation from IR video to EEG modality can be achieved with essentially the same accuracy as the feature fusion version.}
\end{itemize}

\section{Analysis}\label{sec:analysis}

\subsection{Ablation Analysis}
\begin{table}[]
 \small
 \centering
\caption{The complete ablation on loss formulation. Metric:Top1 accuracy (\%).}
\setlength\tabcolsep{0.55pt}
\resizebox{\linewidth}{!}{
\begin{tabular}{cl|lcl|lcl|lcllcllcllcllc}
\hline
\multicolumn{1}{l}{} &  &  & \multicolumn{1}{l}{}        &  &  & \multicolumn{1}{l}{} &  & \multicolumn{14}{c}{Per-class Accuracy}                                                                                                                                                      \\
Methods              &  &  & Module                      &  &  & Accuracy             &  &  & W             &                       &  & N1            &                       &  & N2            &                       &  & N3            &                       &  & REM           \\ \hline
baseline($v_0$)             &  &  & $L_{CE}$                      &  &  & 52.2                 &  &  & 64.7          & \multicolumn{1}{l|}{} &  & 40.3          & \multicolumn{1}{l|}{} &  & 48.4          & \multicolumn{1}{l|}{} &  & 47.8          & \multicolumn{1}{l|}{} &  & 57.3          \\
$v_1$                   &  &  & $L_{CE}$+$L_{C}$+$L_{JSD}$        &  &  & 56.1                 &  &  & 67.7          & \multicolumn{1}{l|}{} &  & 43.8          & \multicolumn{1}{l|}{} &  & 52.0          & \multicolumn{1}{l|}{} &  & 50.4          & \multicolumn{1}{l|}{} &  & 59.2          \\
$v_2$                   &  &  & $L_{CE}$+$L_{D}$+$L_{C}$          &  &  & 62.3                 &  &  & 73.4          & \multicolumn{1}{l|}{} &  & 49.0          & \multicolumn{1}{l|}{} &  & 58.0          & \multicolumn{1}{l|}{} &  & 57.2          & \multicolumn{1}{l|}{} &  & 65.8          \\
$v_3$                   &  &  & $L_{CE}$+$L_{D}$+$L_{JSD}$        &  &  & 60.1                 &  &  & 71.5          & \multicolumn{1}{l|}{} &  & 45.6          & \multicolumn{1}{l|}{} &  & 55.4          & \multicolumn{1}{l|}{} &  & 54.5          & \multicolumn{1}{l|}{} &  & 63.6          \\
$v_4$                   &  &  & $L_{CE}$+$L_{D}$+$L_{JSD}$+$L_{C'}$ &  &  & 61.1                 &  &  & 72.4          & \multicolumn{1}{l|}{} &  & 46.8          & \multicolumn{1}{l|}{} &  & 56.7          & \multicolumn{1}{l|}{} &  & 55.8          & \multicolumn{1}{l|}{} &  & 64.6          \\
ours                 &  &  & \textbf{$L_{CE}$+$L_{D}$+$L_{C}$+$L_{JSD}$}  &  &  & \textbf{64.4}        &  &  & \textbf{75.6} & \multicolumn{1}{l|}{} &  & \textbf{51.0} & \multicolumn{1}{l|}{} &  & \textbf{60.2} & \multicolumn{1}{l|}{} &  & \textbf{59.3} & \multicolumn{1}{l|}{} &  & \textbf{67.8} \\ \hline
    \end{tabular}}
\label{tab: newab}
\end{table}

\begin{figure*}[t]
  \includegraphics[width=\textwidth]{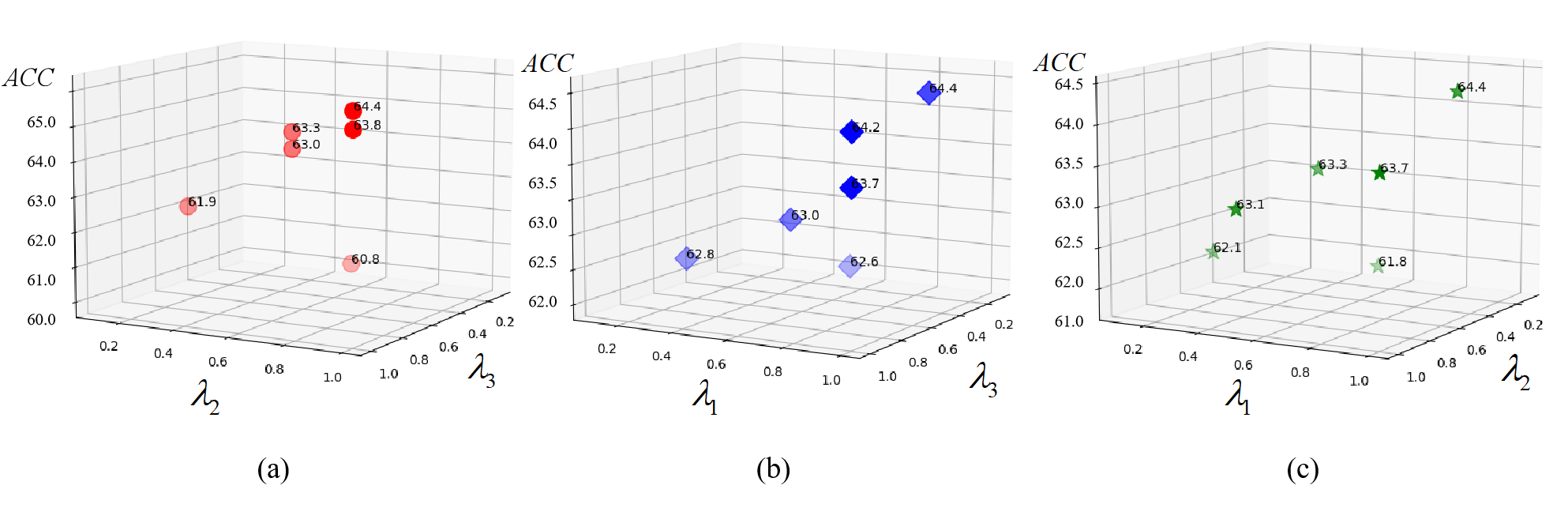}
  \caption{Visualization of hyperparametric analysis. (a) shows the relationship between hyperparameters $\lambda_{2}$ and $\lambda_{3}$ and the overall classification accuracy for the case of fixed $\lambda_{1}$=0.5; (b) shows the relationship between hyperparameters $\lambda_{1}$ and $\lambda_{3}$ and the overall classification accuracy for the case of fixed $\lambda_{2}$=1.0; (c) shows the relationship between hyperparameters $\lambda_{1}$ and $\lambda_{2}$ and the overall classification accuracy for the case of fixed $\lambda_{3}$=1.0.}
  \label{fig: hyper}
\end{figure*}
To demonstrate that each loss function we use is necessary, we remove a particular loss function item by item and observe the change in overall accuracy and per-class accuracy.

As shown in TABLE \ref{tab: newab}, the specific experimental setup is as follows. Experiment $v_0$  removes our proposed $L_C$, $L_D$ and $L_{JSD}$ losses, keeping only the cross-entropy loss $L_{CE}$; Experiment $v_1$ removes the $L_D$ loss compared to the complete SACD;
Experiment $v_2$ removes the $L_{JSD}$ loss compared to the complete SACD; Experiment $v_3$ removes the two-way K-hard negative loss $L_C$ compared to the complete SACD; It is worth noting that the $L_{C'}$ in Experiment $v_4$ represents only a single direction of $L_C$, and this set of experiment is also designed to verify the necessity of the two way $K$-hard Negative loss.
TABLE \ref{tab: newab} shows that Experiment $v_2$ performed significantly worse than the complete SACD, with 2.1\% (64.4\%-62.3\%) worse in overall accuracy and 2.2\% (75.6\%-73.4\%), 2.0\% (51.0\%-49.0\%), 2.2\% (60.2\%-58.0\%), 2.1\% (59.3\%-57.2\%) and 2.0\% (67.8\%-65.8\%). This suggests that loss $L_{JSD}$ is effective and that the differences between categories on $L_{JSD}$ are not significant.
Next, we compare the SACD with two ablation baselines. Experiment $v_4$ and Experiment $v_3$, they are each removed one constraint at a time (one-way $K$-hard Negative loss).
It is easy to observe that the first $K$-hard Negative contrastive learning significantly impacts classification accuracy. 
Taking overall accuracy as an example, when we progressively remove one-way $K$-hard Negative loss and two-way $K$-hard Negative loss, the accuracy degrades 3.3\% (64.4\%-61.1\%) and 4.3\% (64.4\%-60.1\%), respectively. 
It is worth noting that this loss function $L_C$ seems to be very sensitive to different classes as its effects on the $N1$, $N2$, and $N3$ stages are more significant than on the $W$ and $REM$ stages. This confirms the benefit of our structure-aware coarse-grained semantic alignment, and contrastive learning distills more about $N1$, $N2$, and $N3$ stages information from EEG to IR video modality. 
 As shown in experiment $v_1$ in TABLE \ref{tab: newab}, removing the $L_D$ loss function reduces the overall accuracy of SACD by 8.3\% (64.4\% - 56.1\%); the reduction in per-class accuracy is similar to the decrease in overall accuracy (7.2\% to 8.9\% per class). These results indicate that $L_D$, $L_C$, and $L_{JSD}$ are complementary, and they work synergistically to distill knowledge across modalities.

\subsection{Quantitative Analysis} 

\subsubsection{Hyperparameter Analysis}

As described in section \ref{details}, our hyperparameters ($\lambda_{1}$, $\lambda_{2}$ and $\lambda_{3}$) are selected as (0.5, 1, 1), 
the optimal ratio, reaching a overall classification accuracy of 64.4\%, $\lambda_{1}$, $\lambda_{2}$ and $\lambda_{3}$ are the coefficients in front of the loss function $L_D$, $L_C$ and $L_{JSD}$ , respectively.
It is what we have obtained through a lot of comparative experiments. Next, we will give more experimental results about hyperparameters. Based on keeping other experimental settings constant, We set the coefficient of loss$ L_{CE}$ to always be only changed the proportion of ($\lambda_{1}$, $\lambda_{2}$ and $\lambda_{3}$).

To visualize the effect of different hyperparameter ratios on the overall classification performance, as shown in Fig. \ref{fig: hyper}, we plot three figures, each fixing one parameter, to observe the relationship between the change in the other two parameters and the overall accuracy.

Observing Fig. \ref{fig: hyper} (a), we can see that when we fix hyperparameter $\lambda_{1}$ to 0.5 and change the value of ($\lambda_{2}$, $\lambda_{3}$) to (0.5, 1.0), the overall accuracy reach the lowest value of 60.8\% in this group. When ($\lambda_{2}$=1.0, $\lambda_{3}$=1.0), the accuracy get the highest value of 64.4\% in this group of experiments, which shows that when hyperparameter $\lambda_{1}$ is fixed, increasing appropriate hyperparameters $\lambda_{2}$ and $\lambda_{3}$ will improve the contribution to the overall classification accuracy. However, when the two values are rapidly increased to 10, the classification accuracy rapidly decreases to 52\%, which is not drawn due to the view scale, thus showing the importance of the appropriate hyperparameters.

Similar to the first set of experiments, we fix hyperparameter $\lambda_{2}$ to 1.0 and vary different values ($\lambda_{1}$, $\lambda_{3}$) to observe the change in overall accuracy. As shown in Fig. \ref{fig: hyper} (b), when ($\lambda_{1}$=0.5, $\lambda_{3}$=0.1), the overall classification accuracy come to 62.6\%, which is the lowest value in this group of experiments; when ($\lambda_{2}$=0.5,$\lambda_{2}$=1.0), the overall classification accuracy increase to 64.4\%, which is the highest value in this group of experiments. When the value of ($\lambda_{1}$, $\lambda_{3}$) is changed from (0.5, 1.0) to (1, 1), the accuracy decrease by 0.02\% (64.4\%-64.2\%), which also illustrates that the value of hyperparameter $\lambda_{1}$ cannot be increased without a limit.
In Fig. \ref{fig: hyper} (c), we fix the hyperparameter $\lambda_{3}$= 1 and vary the values of hyperparameter $\lambda_{1}$ and hyperparameter $\lambda_{2}$. When the value of ($\lambda_{1}$, $\lambda_{2}$) is (0.5, 0.1), the accuracy comes to 61.8\%, which is the lowest value in this set of experiments, and when the value of ($\lambda_{1}$, $\lambda_{2}$) is (0.5, 1.0), the accuracy rises to 64.4\% which is the highest value in this set of experiments.

\begin{figure*}[t]
  \includegraphics[width=\textwidth]{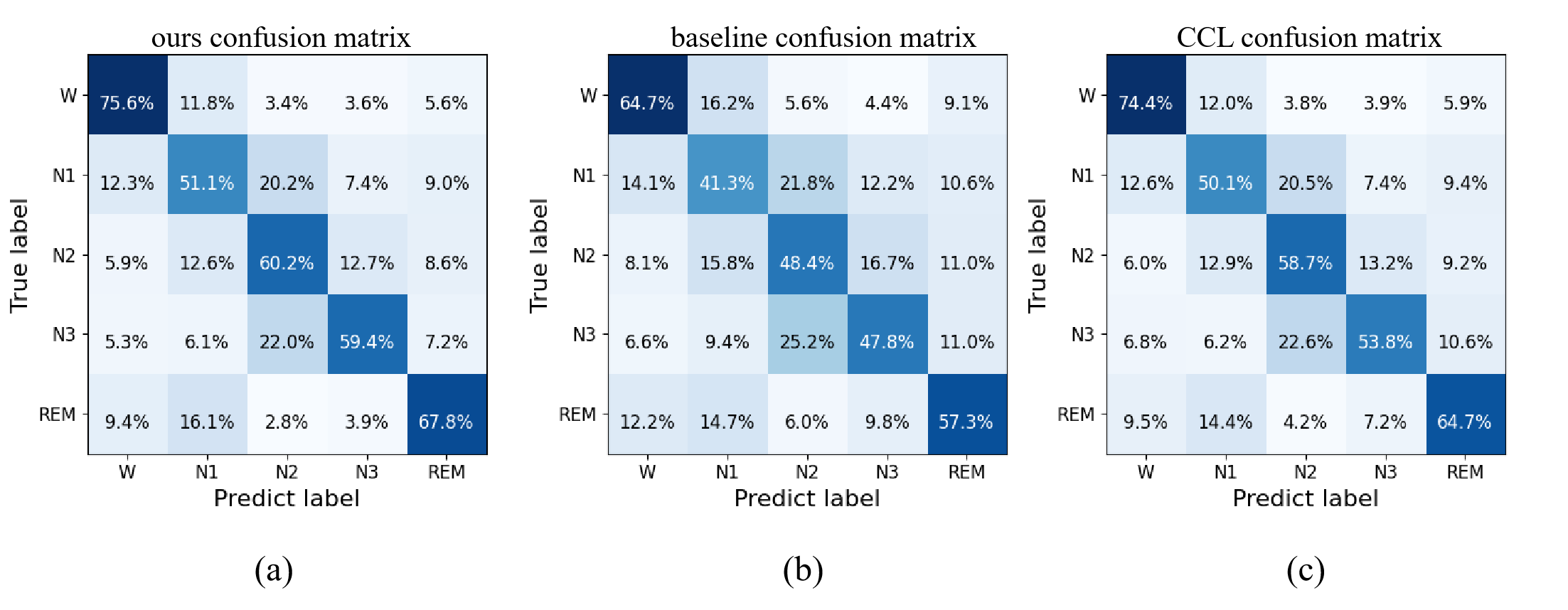}
  \caption{ Normalized confusion matrices of the classification accuracies from our dataset $S^3VE$. (a). confusion matrix for our method SACD. (b).  confusion matrix for baseline R(2+1)D. (c). confusion matrix for CCL \cite{chen2021distilling}. }
  \label{fig: cm}
\end{figure*}

\subsubsection{Confusion Matrix Analysis}
\label{sec: error matrix}

Fig. \ref{fig: cm} shows the results of the confusion matrix predicted by the model. From left to right are (a): our SACD confusion matrix, (b): the IR baseline confusion matrix (c): the CCL confusion matrix. As can be seen with the confusion matrices, the error rate of $N1$ is the highest among all three methods. The poor performance of $N1$ can be attributed to the fact that many samples in $N1$ stage are misclassified as $W$ and $N2$ stages, since most samples in the $N1$ stage belong to the sleep transition period \cite{chen2020sleep} \cite{jia2022multi}. In addition, it is also possible that the number of data in stage $N1$ itself is relatively small compared to the other sleep stages. Also, observing Fig. \ref{fig: cm} (a), it can be found that, except for $REM$, adjacent sleep stages generally tend to have higher confusion rates than other non-adjacent stages. In contrast, $REM$ is more likely to be misclassified as a $W1$ stage. We guess the possible reason is that $N1$ is a light sleep in the immediate sleep stage, and there may be some small erratic movements, such as manual and eye movements.
However, in $REM$, most dreams occur, and there are obvious rapid eye movements. These similar features in the IR video could lead to the occurrence of misclassification. Comparing Fig. \ref{fig: cm} (a) and Fig. \ref{fig: cm} (c), we can see that we have a clear advantage over CCL \cite{chen2021distilling} in distinguishing the two sleep stages, $REM$ and $N3$, which will also be mentioned in the visualization analysis section.


\subsubsection{Statistical Correlation Analysis}
\begin{figure*}[t]
  \centering
  \includegraphics[width=0.8\textwidth]{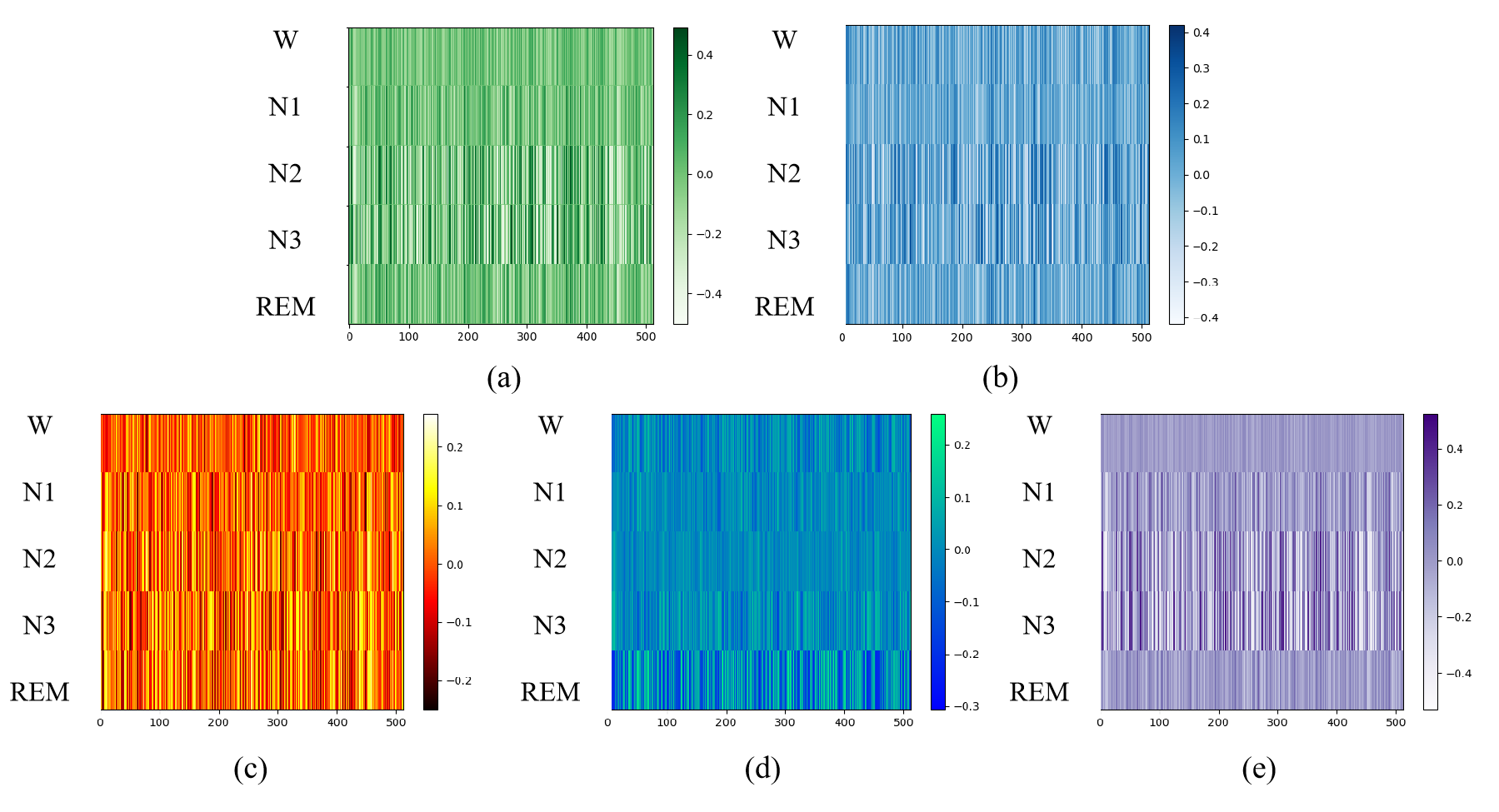}
  \caption{Spearman correlation analysis of SACD's output and AHI. (a) is the Spearman correlation between the ``average condition'' of each individual's five sleep stages and the overall AHI distribution; (b)-(e) is the Spearman correlations between the ``average condition'' of each individual's five sleep stages and the one-hot vector of each AHI class.}
  
  \label{fig: 5cl}
\end{figure*}

\begin{figure}[t]
  \includegraphics[width=0.50\textwidth]{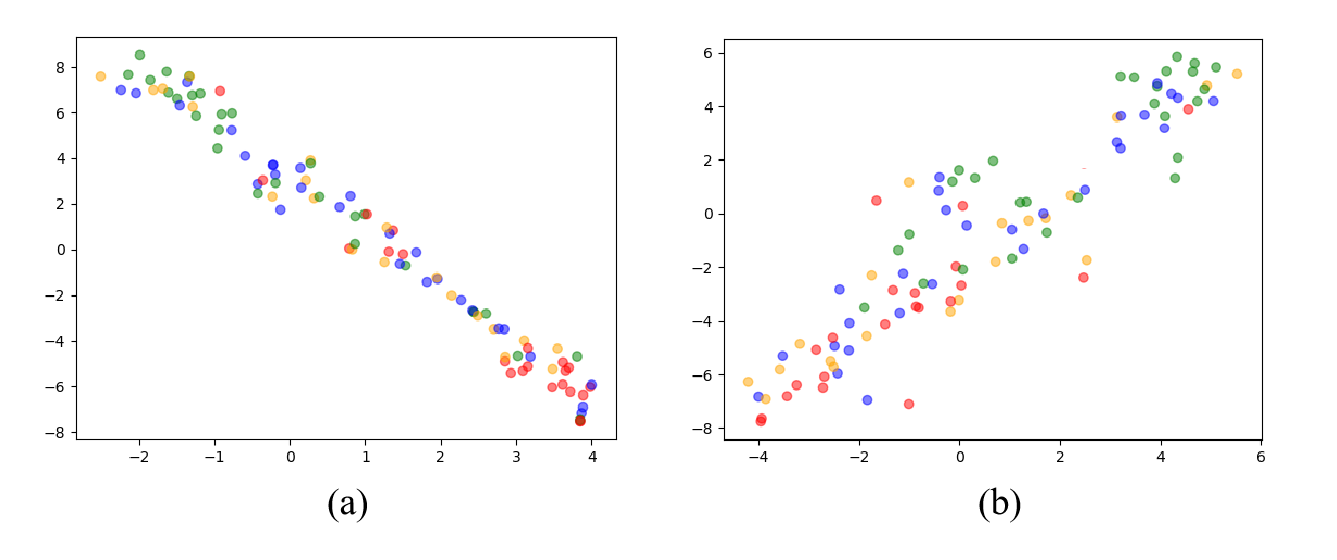}
  \caption{ Selective Tsne visualisation of all samples $N2$ and $N3$. Output channels in $N2$ (Fig. \ref{fig: n2n3} (a)) and $N3$ (Fig. \ref{fig: n2n3} (b)) with AHI speatman correlation coefficients larger than 0.4, green for AHI Normal (AHI \textless 5) , blue for Mild (5\textless=AHI\textless15), orange for Moderate (15\textless=AHI\textless=30), and red for Severe (AHI \textgreater30).}
  \label{fig: n2n3}
\end{figure}

To observe the correlation between the model output and the AHI (interest to sleep apnea physicians), a total of 102 data from the training and test sets were subjected to statistical analysis. The details are as follows: averaging all five sleep stage features output for each individual throughout a night according to the ground truth labels. The averaged results represent the ``average condition'' of the person's five sleep stages, all of which have a dimension of 512 vectors. The AHI was divided into four levels (Normal: \textless 5, Mild: 5-15, Moderate: 16-30, Severe: \textgreater 30) to create an AHI feature distribution, which was adjusted into four one-hot vectors.  Fig. \ref{fig: 5cl} (a) is the Spearman correlation between the ``average condition'' of each individual's five sleep stages and the overall AHI distribution; Fig. \ref{fig: 5cl} (b)-(e) is the Spearman correlations between the ``average condition'' of each individual's five sleep stages and the one-hot vector of each AHI level. The p-value values are all less than 0.05, indicating our statistical results are confident.

As shown in Fig. \ref{fig: 5cl} (a), we can observe that the correlations for $N2$ and $N3$ are significantly higher than those for $W$, $N1$, and $REM$. Consistent with our intuitive sense, there should indeed be a more negligible correlation between the characteristics of $W$ and AHI, as the characteristics of a person while awake are inherently less relevant to sleep apnea events. As shown in Fig. \ref{fig: 5cl} (b) and Fig. \ref{fig: 5cl} (d), similar to the previous analysis, $N2$ and $N3$ are significantly more correlated than $W$, $N1$, and $REM$, suggesting that $N2$ and $N3$ are more closely related to whether they are Normal (AHI \textless 5) or Sever (AHI \textgreater 30). This finding is helpful for clinical medicine, where sleep apnea physicians can look primarily at these two sleep stages. As shown in Fig. \ref{fig: 5cl} (c), the $N2$, $N3$, and $REM$ correlations are more similar than the $W$ and $N1$ stages. As shown in Fig. \ref{fig: 5cl} (d), the correlation between $REM$ is higher than that of the other sleep stages. This finding is inconsistent with our intuition and worthy of further exploration. 

\begin{figure}[t]
  \includegraphics[width=0.50\textwidth]{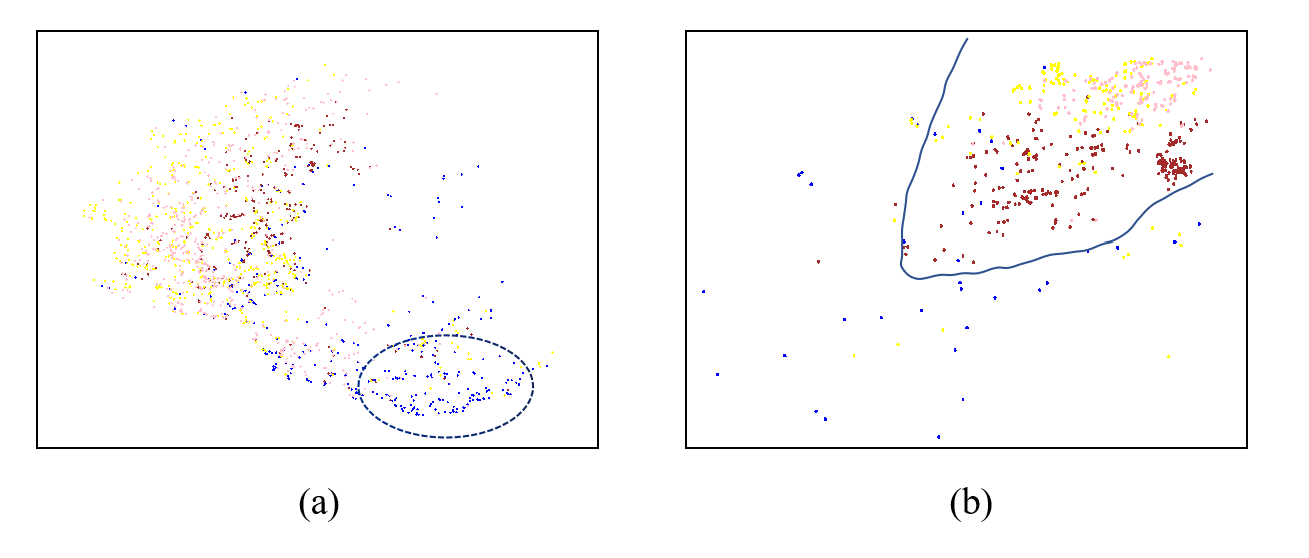}
  \caption{ Tsne visualization of four sample $N2$ and $N3$ clips features. The different colors represent the samples' sleep $N2$ (Fig. \ref{fig: 2li} (a)) or $N3$ (Fig. \ref{fig: 2li} (b)) clips; \textcolor{blue}{Blue represents sample $A$;}  \textcolor{pink}{ Pink represents sample $B$;} 
  \textcolor{yellow}{ Yellow represents sample $C$;}  \textcolor{brown}{ Brown represents sample $D$}; their basic information is shown in the TABLE \ref{tab：ABCD}, and we have also marked  the ``danger zones'' where the AHI is larger.}
  \label{fig: 2li}
\end{figure}

We visualize the output features from the following two perspectives to further visualize the correlation between the model output features and the AHI. Fig. \ref{fig: n2n3}: Tsne \cite{van2008visualizing} visualization of the high weight channel for the average feature of all samples $N2$ and $N3$, respectively, and Fig. \ref{fig: 2li}: tsne visualization of the $N2$ and $N3$ clips (30s) features for the four samples.
As shown in Fig. 8, we make the selective output of the $N2$ and $N3$ feature channels in Fig. \ref{fig: 5cl} (a) that are highly correlated with AHI. It is done by selecting all channels with a spearman correlation greater than 0.4, which is also a selective dimensionality reduction process. For $N2$, there are 17 such channels; for $N3$, there are 36 such channels; finally, Visualization by Tsne, respectively. The different colors in Fig. \ref{fig: n2n3} indicate different AHI levels, and we can observe that in (a), the normal samples (green dots) are primarily located at the top left of the figure, and the severe samples (red dots) are primarily located at the bottom right of the figure. It can be more clearly seen that certain features of the $N2$ output are correlated with the AHI. Similarly, in Fig. \ref{fig: n2n3} (b), the most severe samples are located at the bottom left of the figure, and the most normal samples are located at the top right of the figure, which leads to a similar conclusion.

While the Fig. \ref{fig: 5cl} visualizations are performed on the average features of all samples in the sleep stages, we next chose four more representative samples and do Tsne \cite{van2008visualizing} visualizations of all their $N2$ and $N3$ clips. As shown in Fig. \ref{fig: 2li}, unlike in Fig. \ref{fig: n2n3}: the different colors in the figure represent different samples, and each dot represents the 512-dimensional feature distribution of a single clip for one sample. We also give the basic information of the four individuals as a reference, as shown in the TABLE \ref{tab：ABCD}. In Fig. \ref{fig: 2li} (a), it can be observed that the larger the AHI, the more clips of the sample are distributed in the lower right of the plot, especially the blue dots, with a high AHI of 94.2 TST. A similar conclusion can be drawn in Fig. \ref{fig: 2li} (b), where the larger the AHI, the more clips of the sample are distributed in the lower left of the plot. These two \textbf{``danger zones''} (areas of larger AHI) are drawn in the diagram. Given a new subject, we can do a \textbf{``feature portrait''} in the Tsne graph to get a rough guess of which AHI levels it belongs to.
We have discussed with sleep apnea physicians the feasibility of this kind of \textbf{``feature portrait''} in practical clinical application and have validated it on a number of new examples.

\begin{table}[]

 \small
 \centering
\caption{Message of samples ABCD ;
TST : Total Sleep Time.\\(Normal:  \textless 5, Mild : 5-15, Moderate: 15-30, Severe: \textgreater 30)}
\setlength\tabcolsep{0.55pt}
\resizebox{\linewidth}{!}{
\begin{tabular}{c|c|c|c|c|c}
\hline
Index & Data     & Gender & Age & Weight & \begin{tabular}[c]{@{}c@{}}Sleep-related apnea-hypopnea Index\\       (AHI)\end{tabular} \\ \hline
A     & 20210310 & Male   & 37  & 108kg  & 94.2 TST                                                                                  \\ \hline
B     & 20200117 & Male   & 31  & 66kg   & 0.4 TST                                                                                   \\ \hline
C     & 20210326 & Female & 58  & 58kg   & 6.9 TST                                                                                   \\ \hline
D     & 20210329 & Female & 65  & 64kg   & 26.2 TST                                                                                  \\ \hline
\end{tabular}}
\label{tab：ABCD}
\end{table}
\subsection{Qualitative Results} 

\begin{figure}[t]
  \includegraphics[width=0.49\textwidth]{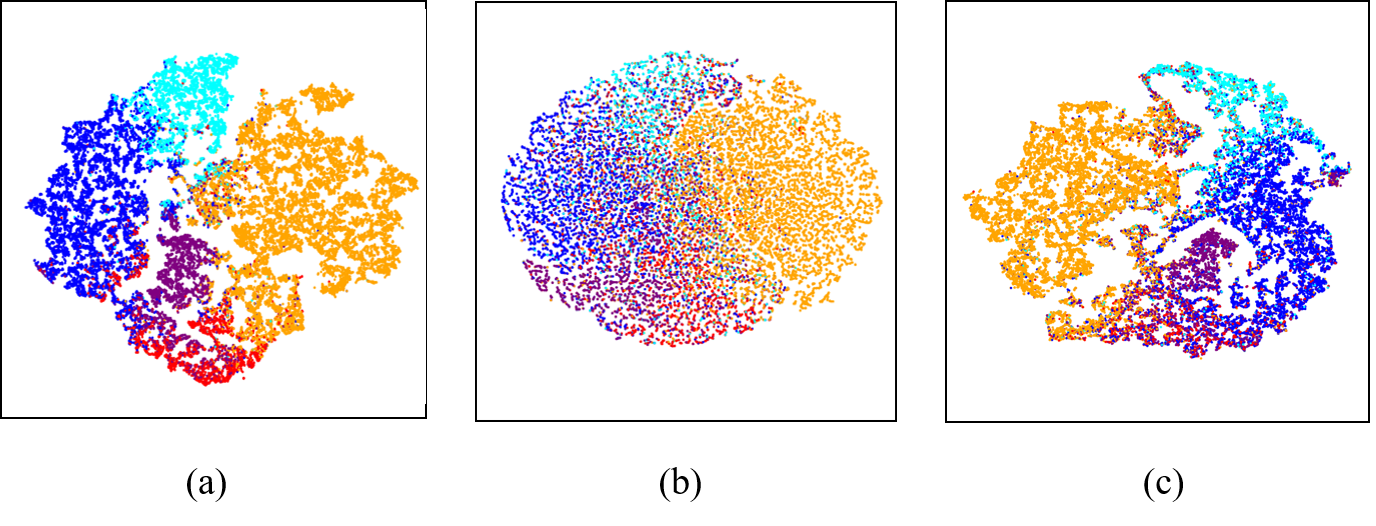}
  \caption{Visualisation with Tsne \cite{van2008visualizing}, (a) is the feature embedding extracted from $S^3VE$ test set using our method. (b) is the feature embedding extracted from $S^3VE$ using baseline Res(2+1)D. (c) is the feature embedding extracted from $S^3VE$ using CCL \cite{chen2021distilling} (the orange represents $W$; cyan represents $N1$; blue represents $N2$; red represents $N3$; purple represents $REM$).}
  \label{fig: tsne}
\end{figure}


\subsubsection{Visualisation Analysis}
To qualitatively understand video representation and cross distillation, we analyze the SACD with visualizations.
When using Tsne \cite{van2008visualizing} to visualize the 512-dimensional embeddings of the different sleep stages, we can see that embeddings of different categories are grouped into different clusters. The distances between dots in Fig. \ref{fig: tsne} represent the distances of embeddings in the higher dimensional space, and one can observe that the classification results make the distances between identical classes (sleep stages) closer and the distances between different classes (sleep stages) farther, thus showing the success of our method. Compared with the aggregation cluster of the baseline (Fig. \ref{fig: tsne} (b)) samples, the cluster of each class of our SACD (Fig. \ref{fig: tsne} (a)) is significantly more concentrated, which illustrates that our method SACD's embeddings from the same class have a higher similarity. Compared to the aggregated clusters of CCL \cite{chen2021distilling} (Fig. \ref{fig: tsne}  (c)), there are significantly fewer overlapping dots of different colors, implying that our embeddings are more accurate; specifically, the purple ($REM$) and red ($N2$) parts of Fig. \ref{fig: tsne} (c) are more confounded, whereas the species in Fig. \ref{fig: tsne} (a) are significantly more clearly separated, indicating that our method classifies better in $REM$ and $N3$ and this conclusion is corroborated by the section \ref{sec: error matrix} confusion matrix analysis. In summary, the qualitative results show that our SACD can learn discriminative video representations of cross-modal distillation from EEG modality and that the distillation performance is better than current state-of-the-art distillation methods.

\begin{figure*}[t]
  \includegraphics[width=1\textwidth]{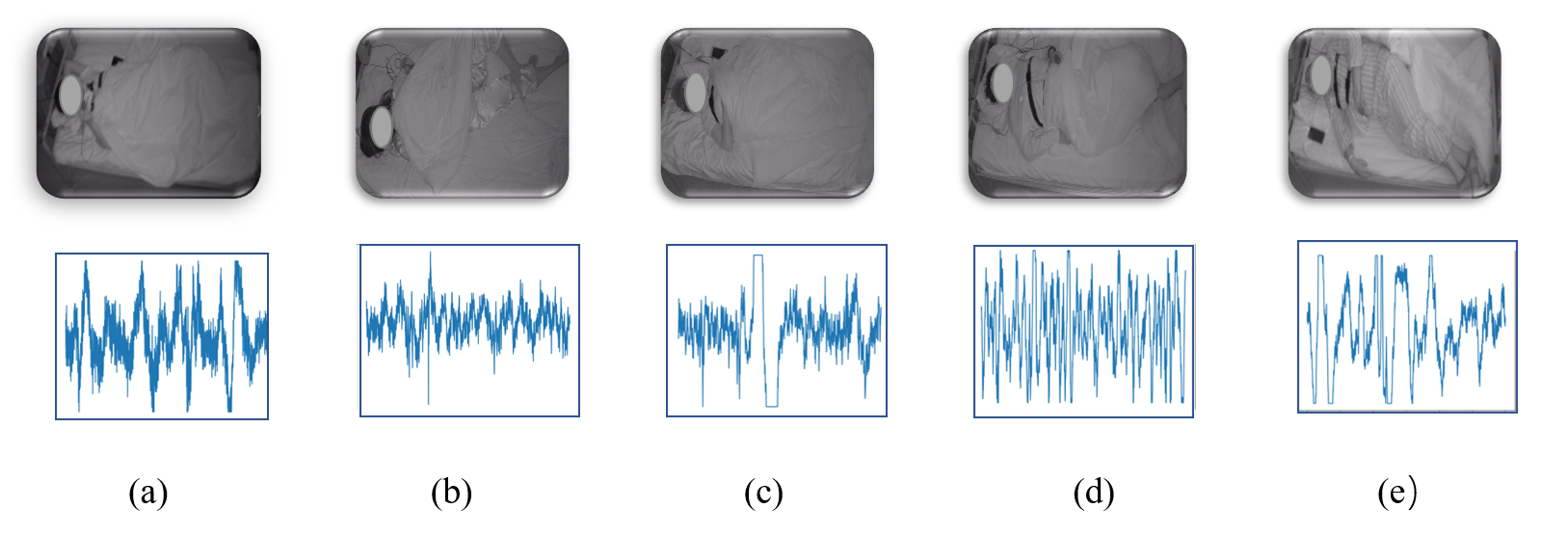}
  \caption{ IR video-EEG correspondence:visual representation of five sleep stages and their EEG waves. In these examples, our predictions are correct and the baseline predictions are wrong. (a) a clip labeled by $W$ . (b) a clip labeled by $W$ . (c) a clip labeled by $N2$ . (d) a clip labeled by $N3$. (e) a clip labeled by $REM$. }
  \label{fig: 5class}
\end{figure*}

\begin{figure*}[h]
  \includegraphics[width=1\textwidth ]{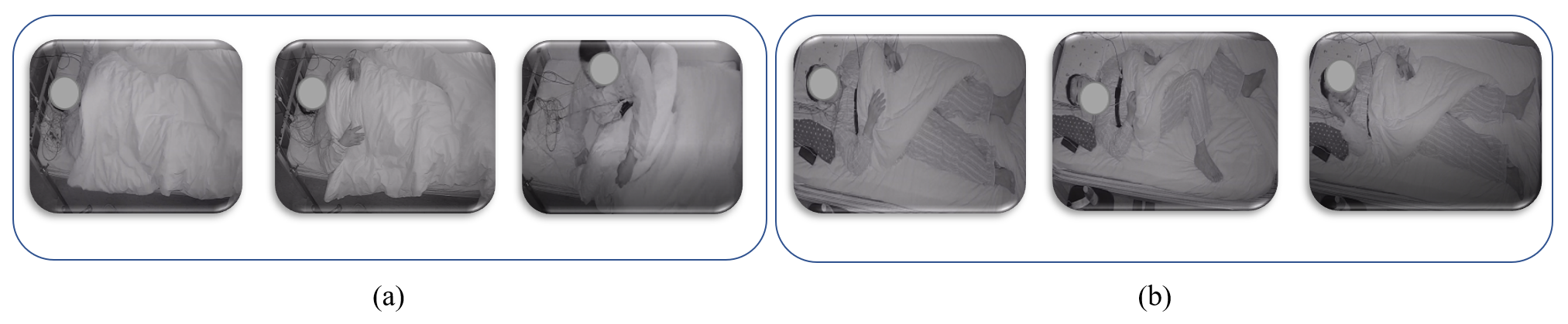}
  \caption{ Bad case examples: (a) example of sleep stage junction, previous clip is $N3$ and next clip is $W$. (b) Instance-level specificity, this instance's movement changes more frequently than normal instance}
  \label{fig: wrong}
\end{figure*}

\begin{figure*}
  \centerline{\includegraphics[width=\textwidth]{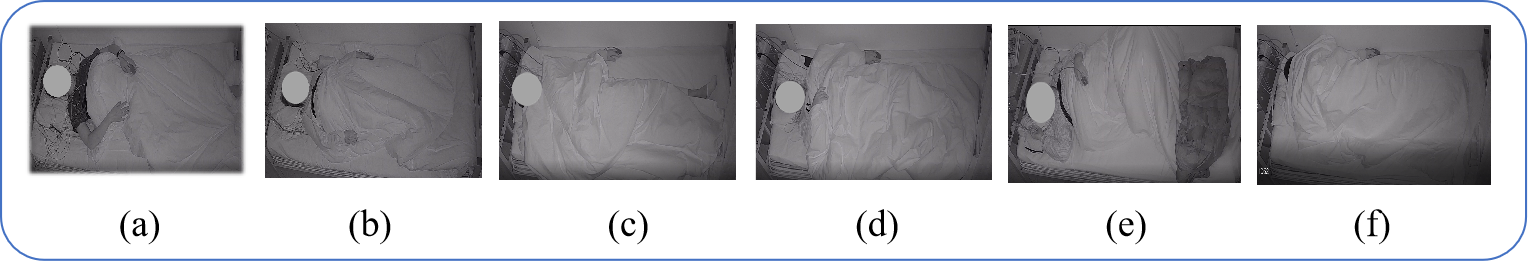}}
  \caption{ Analysis of the main error types. (a) The most common error in the stage $W$, wrongly evaluated as stage $N1$;
  (b) The most common error in the stage $N1$, wrongly evaluated as stage $N2$;
  (c) The most common error in the stage $N2$, wrongly evaluated as stage $N1$;
  (d) The most common error in the stage $N3$, wrongly evaluated as stage $N2$;
  (e) coexisting errors;\\
  }
  \label{fig: error}
\end{figure*}

\subsection{Qualitative Analysis on Cross-Modal Correspondence}
To understand the cross-modal semantic correspondence clearly, we provide examples of video-EEG correspondence in Figure \ref{fig: 5class}. We select five videos of five sleep stages for analysis and gave the EEG waveforms of these five videos to observe together. The five sleep infrared videos are misclassified using baseline R(2+1)D, but after our SACD distillation, the results are correct. First, looking at the EEG waveform figure, we can see that in Fig. \ref{fig: 5class} (a), the EEG waveform is a $beta$ wave, which usually appears in $W$ \cite{sleep2001proposed}. In Fig. \ref{fig: 5class} (c), we can see the $delta$ wave, which usually appears in $N2$ \cite{sleep2001proposed}, and in Fig. \ref{fig: 5class} (e), the we can see the $theta$ wave, which usually appears in $REM$ \cite{sleep2001proposed}. This waveform information is easily determined by observing the EEG, indicating that the EEG is the discernible modality with a prior knowledge for these examples. And these information cannot be used without the cross-modal distillation, resulting in ``prediction drift'' that is prone to occur without distillation, such as baseline R(2+1)D, and the results after distillation are all correct. This demonstrates the usefulness of the EEG modality for our distillation task and the effectiveness of our method.

In addition to these correct examples, we also find some unsolvable examples in the experiment. As shown in Fig. \ref{fig: wrong} (a), both the baseline and our method have been misjudged. We guess that the reason may be that these videos may be at the critical point of sleep stage change. This speculation has been mentioned in a number of previous studies \cite{chen2020sleep}\cite{jia2022multi}.
It is reflected at the embedding level that the clustering of feature embeddings may not be close enough. Our other speculation is that the error is caused by the PSG labeling process: the PSG gives a unique sleep classification for each clip (30s) based on a combination of physiological signals, with the potential problem that if the clips are partly in one sleep stage and partly in another, then the PSG will give the final label based on their proportion of the sleep stage. This indirectly leads to less accurate labeling, especially when the sleep stage is switched, and is more likely to be misclassified. However, this problem is inherent to PSG and has nothing to do with our algorithm, and we will find ways to correct this problem in future research.
In Fig. \ref{fig: wrong} (b), this is a special instance that frequently moved throughout sleep. It gives us the heuristic that compares to class-level. We should probably pay more attention to the instance level when faced with these special cases. In addition, sleep is a long-term problem; both our method SACD and baseline do not take advantage of the influence of time correlation, the correlation on the time axis may help us a lot in $S^3VE$, which is also our direction for future work.

We produce the confusion matrice in \ref{sec: error matrix} for our method SACD on the test set of $S^3VE$ to observe the most dominant error types for each sleep stage. According to the confusion matrix, we list the main error types corresponding to the five stages, as shown in  Fig. \ref{fig: error} .
About Fig. \ref{fig: error} (a), in stage $W$, the most dominant error is being misclassified as $N1$; We believe that there may be some sleep-onset phases at the end of $W$ and the beginning of $N1$, where physical representations are remarkably similar and complex to distinguish.
About Fig. \ref{fig: error} (b), in stage $N1$, the most dominant error is being misclassified as $N2$;
About Fig. \ref{fig: error} (c), in stage $N2$, the most significant errors are misclassified as $N1$ and $N3$, and here we have chosen the example of misclassified N1 for illustration; It may be that the proportion of N1 instances is relatively tiny, and it is difficult to make better use of contrastive learning to widen the gap between $N2$ and $N1$. This is a problem of sample imbalance. Nevertheless, we expect to continue to collect some $N1$ data or use data enhancement methods in the future.
About Fig. \ref{fig:  error} (d), in stage $N3$, the most dominant error is being misclassified as $N2$; Similar to Fig. \ref{fig: error} (a), this error belongs to the problem that the intersection time is difficult to define.
About Fig. \ref{fig: error} (e), in stage $REM$, the most dominant error is being misclassified as $N1$; This error is more interesting. N1 belongs to the light sleep stage, which is more into sleep. Some samples will show that the body and expression are not completely relaxed, which is similar to the rapid eye movement in $REM$ stage and the body movements during dreaming—leading to misjudgment.
 
About Fig. \ref{fig: error} (f), this is a possible problem in any sleep stage because the samples sometimes turn their heads to the side of the wall, and the infrared camera cannot capture the face; sometimes, the samples even cover their heads with quilts. We are also unable to obtain more information; the solution is to install an infrared camera on the edge just above the wall to get infrared video information from multiple viewing angles.


\section{Conclusions}
In this paper, we propose a novel cross-modal distillation dataset and benchmark for the multi-modal community. This dataset bridges the gap between the clinical and the visual modality and promotes the developments of the point-of-care research. Besides the contribution of the dataset, we also present a novel cross-modal distillation method that can effectively reduce the cross-modal gap and facilitate the usage of visual modality to classify the sleep stage. Experimental results show that our method outperforms other SOTA methods in both our dataset and the other benchmarks. With IR video alone, the proposed method can also achieve considerable sleep stage classification performance. We expect the proposed method to be applied in a wider range of scenarios and hope that more researchers will pay attention to the infrared sleep video modality.



\bibliographystyle{IEEEtran}
\bibliography{samples/ref.bib}

\end{document}